\documentclass[journal]{IEEEtran}
\usepackage{amsmath,graphicx,cite,epsfig}
\usepackage{multirow}
\usepackage{booktabs}
\usepackage{float}
\usepackage{subfigure}
\usepackage{caption}
\usepackage{threeparttable}
\usepackage{diagbox}
\usepackage{url}
\usepackage{bm}
\usepackage{algorithm}
\usepackage{algorithmicx}
\usepackage{algpseudocode}
\usepackage{comment}
\usepackage{makecell}
\usepackage{bm}
\usepackage[colorlinks,urlcolor=magenta]{hyperref}
\usepackage{amsfonts}
\usepackage{bbm}

\usepackage[table,xcdraw]{xcolor}
\newcommand{\tcb}{\textcolor{black}}
\newcommand{\tco}{\textcolor{black}}
\newcommand{\tcrt}{\textcolor{black}}

\definecolor{mygray}{gray}{.9}
\ifCLASSINFOpdf

\else

\fi

\begin{document}
%
\title{Semantic-aware Message Broadcasting for Efficient Unsupervised Domain Adaptation}
%
%

\author{Xin Li,~\IEEEmembership{Student Member, IEEE}, Cuiling Lan, Guoqiang Wei,  and  Zhibo Chen,~\IEEEmembership{Senior~Member,~IEEE}
\thanks{This work was supported in part by NSFC under Grant U1908209, 61632001 and the National Key Research and Development Program of China 2018AAA0101400. (Corresponding authors: Zhibo Chen and Cuiling Lan.)}
\thanks{X. Li, G. Wei and Z. Chen are with  University of Science and Technology of China, Hefei 230027, China (e-mail: lixin666@mail.ustc.edu.cn; wgq7441@mail.ustc.eedu.cn;  chenzhibo@ustc.edu.cn). Cuiling Lan is with Microsoft Research Asia (MSRA) (e-mail: culan@microsoft.com)}
}  

\newcommand{\ieno}{\textit{i.e.}}
\newcommand{\egno}{\textit{e.g.}}
\newcommand{\etal}{\textit{et al.}}
\newcommand{\tcr}{\textcolor{}}
\newcommand{\etcno}{\textit{etc}}

\maketitle

\begin{abstract}
Vision transformer has demonstrated great potential in abundant vision tasks. 
However, it also inevitably suffers from poor generalization capability when the distribution shift occurs in testing (\ieno, out-of-distribution data). To mitigate this issue, we propose a novel method, Semantic-aware Message Broadcasting (SAMB), 
which enables more informative and flexible feature alignment \tcb{for unsupervised domain adaptation (UDA)}. 
Particularly, we study the attention module in the vision transformer and notice that the alignment space using one global class token lacks enough flexibility, where it interacts information with all image tokens in the same manner but ignores the rich semantics of different regions. 
In this paper, we aim to improve the richness of \tcb{the} alignment \tcb{features by enabling semantic-aware adaptive message broadcasting}. 
Particularly, we introduce a group of learned \tcb{group} tokens as nodes to aggregate the global information from all image tokens, 
but encourage different group tokens to adaptively focus on the message broadcasting \tcb{to} different semantic regions. 
In this way, \tcb{our message broadcasting encourages} the group tokens to learn more informative and diverse information for \tcb{effective} domain alignment.
Moreover, we systematically study the effects of adversarial-based feature alignment (ADA) and pseudo-label based self-training (PST) \tcb{on UDA}. 
We find that one simple two-stage training strategy with the cooperation of ADA and PST can further improve the \tcb{adaptation capability} of the vision transformer.
Extensive experiments on DomainNet, OfficeHome, and VisDA-2017 demonstrate the effectiveness of our methods for \tcb{UDA}. Our code will  be available at \url{https://github.com/lixinustc/SAMB-TransformerUDA}.
\end{abstract}


%
\IEEEpeerreviewmaketitle

\section{Introduction}
\IEEEPARstart{D}{eep} neural networks (DNNs) have been greatly \tcb{advanced} in recent years in a variety of vision tasks~\cite{dosovitskiy2020imageViT,carion2020endDETR,he2017maskMASKRCNN,redmon2016youYOLO,cheng2022masked,li2022exploring,chu2021twinsTwins}, \egno, classification, segmentation, and object detection. 
However, it still suffers from poor \tcb{performance} when the testing data violate the assumption of identical independently distributed (\ieno, \textit{i.i.d.})~\cite{zhou2022domainSurveyKaiyang,wang2022generalizingSurveyJingdong}. This prevents the application of DNNs in many practical scenarios, where target data are most likely intervened by unknown confounders~\cite{li2021confounderCICF, zhang2022GE-ViTs}, \egno, brightness, background, shape, \etcno. A na\"ive solution for this is to annotate the samples of target data \tcb{for training} but \tcb{this} is labor-intensive and time-consuming. \tcb{As an alternative}, Unsupervised Domain Adaptation \tco{(UDA)} aims to eliminate the domain shift and transfers the knowledge learned from the labeled source domain to the unlabeled target domain, which receives great attention. 

Over the past decade, a spectrum of UDA works~\cite{ganin2016domain_dann,borgwardt2006integratingMMD,ren2022multi,feng2021complementary,moon2022multistage,bai2021hierarchical}  have been \tcb{investigated} based on conventional convolution networks, \egno, ResNet~\cite{he2016deepResNet}, and AlexNet~\cite{krizhevsky2017imagenetAlexNet}. The commonly-used methods can be roughly divided into two categories, distribution alignment~\cite{borgwardt2006integratingMMD,wei2021toalign,tzeng2017adversarialADDA,ganin2016domain_dann}, and \tcb{pseudo-labeling based methods}
~\cite{saito2017asymmetric,zou2019confidenceCRST,gu2020spherical}. In particular, adversarial learning-based distribution alignment has been prominent and popular since the pioneering work DANN~\cite{ganin2016domain_dann}. A discriminator is used to distinguish the source and target domains, which enforces the generator (\ieno, feature extractor) to learn the domain-invariant knowledge in an adversarial way. In contrast, \tcb{pseudo-label based methods learn} the data structure and discriminative information of the target domain with pseudo-labels. However, unreliable pseudo-labels inevitably cause the adverse effects for DNNs due to the domain shift. \tcb{Most of these methods are studies based on CNNs. Whether there is new insight on top of the powerful transformer architecture is still under-explored. }

Recently, Vision Transformers~\cite{dosovitskiy2020imageViT,liu2021Swin,chu2021twinsTwins,wang2021pyramidPVT,xu2022groupvit,zhu2020deformable,cheng2022masked,li2022hst,liu2022swiniqa,lu2022rtn} have been explored in various vision tasks and shown the great potential for their strong \tcb{modeling} capability especially trained with extremely large-scale data. 
Even though it has been experimentally proven that the transformer owns a better generalization ability than CNNs~\cite{xu2021cdtransCDTrans,zhang2022GE-ViTs}, the promising performance of the transformer is also hindered by the domain shift between training and testing data. To tackle this challenge, some works~\cite{yang2021tvtTVT, zhang2022delvingGE-ViTs, sun2022safeSSRT, hoyer2022daformerDAformer} take a step forward and investigate the transformer-based UDA. These works reveal that the distribution alignment~\cite{sun2022safeSSRT,yang2021tvtTVT} and self-training~\cite{xu2021cdtransCDTrans,wang2022domainBCAT} \tcb{can still improve the UDA performance on the} transformer. 
Moreover, self-training in transformer behaviors more favorable and reliable 
than \tcb{that} in CNNs. There are several intrinsic characteristics of the transformer in UDA~\cite{zhang2022GE-ViTs,sun2022safeSSRT,hoyer2022daformerDAformer,xu2021cdtransCDTrans}. 1) From the training perspective, the vision transformer is susceptible to model collapse due to its strong representation capability, which increases the risk of over-fitting to the source domain~\cite{zhang2022GE-ViTs}. This poses a higher requirement for \tco{effective} training strategy. 2) The Vision Transformer mostly focuses on the global contextual information, where one class token is used to interact with all image tokens in the attention module~\tcb{\cite{dosovitskiy2020imageViT}}.
Alignment \tcb{based on} such class token lacks \tcb{flexibility and expressiveness}, which limits the \tcb{efficiency of knowledge transfer} in unsupervised domain adaptation. Moreover, this challenge is seldom explored in existing Transformer-based UDA works.

In this paper, we aim to boost the domain adaptability of the vision transformer \tcb{by} improving the richness and flexibility of the alignment space from the message passing \tcb{perspective}.
 It is noteworthy that self-attention is a basic and crucial component in the vision transformers, which treats each image token (\ieno, a local region) as a ``word", and then, enhances them by modeling their dependencies~\cite{vaswani2017attentionisallyouneed,dosovitskiy2020imageViT}. There are two typical message-passing processes in the self-attention module. 1) Message Aggregation: the class token (one token used for classification) aggregates the global information from all image tokens. 
 2) Message Broadcasting: the message of the class token is distributed to different image tokens in the same manner.
We pinpoint that the unified message-passing process enables the learning of the global contextual information \tcb{but} ignores the semantic flexibility and diversity of the classification space, which limits the domain adaptability of ViT~\cite{dosovitskiy2020imageViT}.
 
Motivated by this, we propose the novel Semantic-adaptive Message Broadcasting (SAMB)
for transformer-based UDA, which enables more informative and flexible alignment space. 
We achieve this by \tcb{introducing} a group of learned \tcb{(group)} tokens as nodes, each of which is responsible for \tcb{broadcasting its information (collected from all image tokens) to the corresponding related semantic regions}.
In this way, the SAMB enforces the group tokens to learn more \tcb{rich} and flexible \tcb{information} for the alignment between different domains. To \tcb{encourage} different group tokens focus on the message passing \tcb{for} different semantics in one image, we \tcb{introduce} the dynamic assignment for each node (group token) \tcb{for ``semantic-aware" information broadcasting}. Specifically, we introduce the Gumbel Softmax~\cite{jang2016categoricalGumbelsoftmax} \tcb{for} our assignment. 

\tcb{In addition, p}revious works have verified the effectiveness of \textbf{Ad}versarial Feature \textbf{A}lignment (ADA) and \textbf{P}seudo-label based \textbf{S}elf-\textbf{T}raining (PST) on Transformer-based UDA. However, the cooperation of ADA and PST has not been explored in Transformer-based UDA. In this paper, we systematically investigate the cooperation strategy for ADA and PST. We obtain the following \tcb{observations/}conclusions. 
 1) Model collapses are more likely to occur when the PST and ADA are used to optimize the network at the same time in a simplified manner. 2) It can further boost the adaptation ability by simply dividing the training process into two stages, where the first stage is optimized by ADA, and the second one uses PST. The reason for that is a stronger \tcb{``starter"} can provide more accurate pseudo-labels for PST.
  3) Jointly optimizing ADA and PST at the second stage is optimal, since the risk of model collapse is reduced at the first stage by ADA.

We validate the effectiveness of our method on image classification tasks (Unsupervised domain adaptation). It is noteworthy that our method is complementary to various transformer-based UDA methods and is easy to be integrated into them. Extensive experiments have shown its applicability and strengths on commonly-used UDA benchmarks, \ieno, DomainNet~\cite{peng2019momentDomainNet}, OfficeHome~\cite{venkateswara2017deepOffice-Home}, and VisDA-2017~\cite{peng2017visdaViSDA2017}. 

 The contributions of this paper can be summarized as follows:
 \begin{itemize}
     \item We propose Semantic-aware Message Broadcasting (SAMB) to improve the domain adaptability of vision transformers. Concretely, we dynamically assign  different group tokens for the message broadcasting of different semantics in the same image, which enforces the group tokens to learn more informative and diverse knowledge for the domain alignment.  

     \item We systematically investigate the cooperation strategies of adversarial feature alignment and pseudo-label based self-training for Transformer-based UDA. Experimental observation provides a simple but effective two-stage training mechanism for Transformer-based UDA, which improves the adaptation ability while eliminating the model collapse. 

     \item Extensive experiments on image classification have validated the effectiveness and applicability of our proposed method on Transformer-based UDA. 
     
 \end{itemize}

\begin{figure*}
    \centering
    \includegraphics[width=0.95\linewidth]{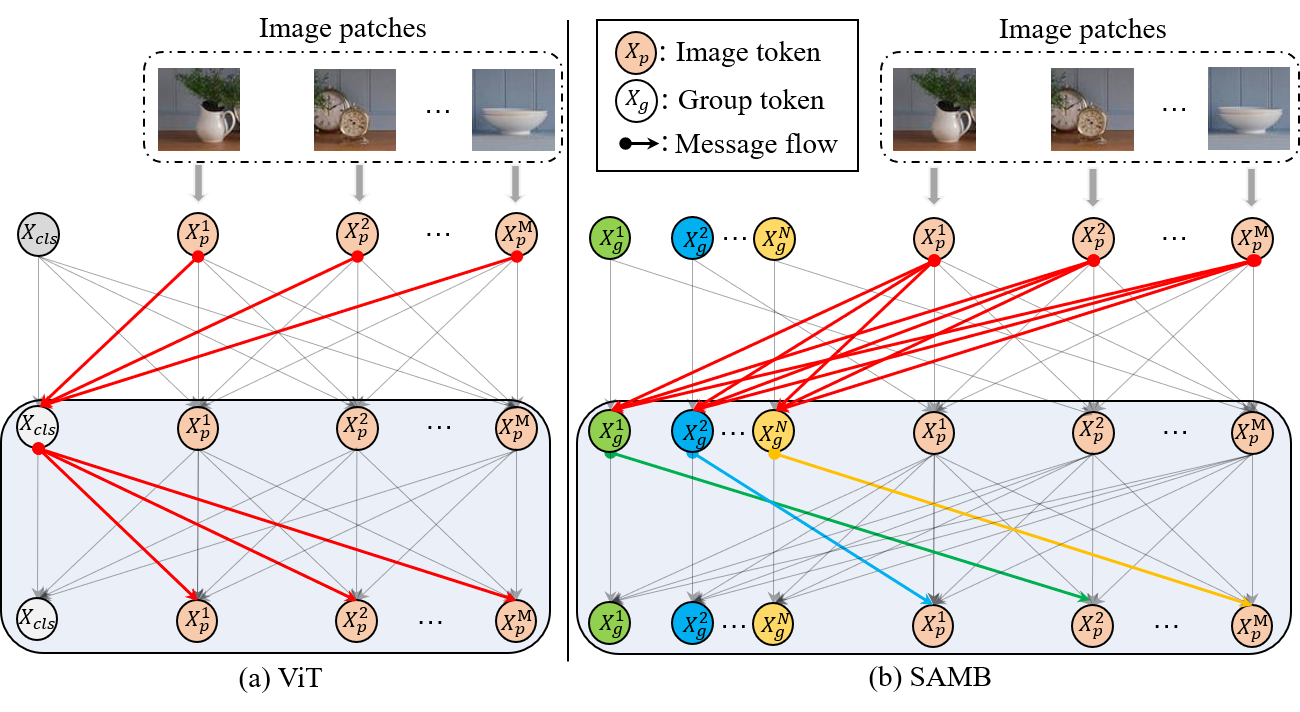}
    \caption{\tcb{We illustrate the information flow in a self-attention layer. The first row to the second row illustrates the aggregation of information to each given token while the second row to the third row illustrates the broadcasting of information from each given token to other tokens.} A comparison between (a) previous message passing \tcb{with information broadcasting from class token $X_{cls}$} in the self-attention layer of vision transformer, and (b) our proposed spatial-adaptive message passing \tcb{with semantic-aware information broadcasting from \emph{group} tokens $X_{g}^k, k=1,\cdots,N$}. $X_p^i, i=1, \cdots, M$ denote the image tokens.} 
    \label{fig:concept_comparion}
\end{figure*}

The remaining parts of this paper are organized as follows. We describe the related works in Section \ref{sec:related work}. In Section \ref{sec:methods}, we first introduce the background for unsupervised domain adaptation and vision transformer. Then, we describe our proposed \tcb{semantic-aware} message broadcasting and \tcb{joint} training mechanism {of adversarial feature alignment and pseudo-label based self-training} in detail. The experimental results and analysis are presented in Sec.~\ref{sec:experiments}. Finally, we present the conclusion in Sec.~\ref{sec:conlusion}.
Our codes will be released upon acceptance. 

\section{Related works}
\label{sec:related work}
\subsection{Unsupervised Domain Adaptation}
Unsupervised Domain Adaptation (UDA)~\cite{ganin2016domain_dann,borgwardt2006integratingMMD,xu2021cdtransCDTrans,sun2022safeSSRT,deng2021joint,xu2021neutral,xu2022few,dai2021disentangling,lu2022styleam,liu2022source} aims to transfer learned knowledge from the labeled source domain(s) to the unlabeled target domain. Previous works on UDA 
can be roughly divided into two categories, \ieno, distribution alignment, and self-training, respectively. Early existing alignment-based methods tend to learn domain-invariant representation by explicitly reducing the domain discrepancy, which is measured by some distribution discrepancy metrics~\cite{borgwardt2006integratingMMD, zellinger2017centralCMD, sun2016returnCORAL, sun2016deepdeepcoral, peng2019momentM3SDA}. 
Recently, adversarial learning-based alignment has shown overwhelming advantages in the unsupervised domain adaptation since the pioneering work DANN~\cite{ganin2016domain_dann}, where a domain discriminator is used to distinguish the source and target domains in the feature space. Meanwhile, a generator (\ieno, feature extractor) is trained to extract the domain-invariant representation by fooling the domain discriminator in an adversarial manner. With the advancement of deep learning, abundant excellent variants~\cite{zhang2019domainSymnets, tzeng2017adversarialADDA, hoffman2018cycada, russo2018sourceSBADA, long2018conditionalCDAN, saito2018maximumMCD, xie2018learningMSTN, wang2019transferableTADA, wei2021toalign, chen2022reusingDALN, chen2019progressivePFAN, li2021biBCDM, luo2020unsupervised, chang2019domainDSBN} of adversarial learning-based alignment have been further developed. ADDA~\cite{tzeng2017adversarialADDA} 
combines the discriminative feature learning with untied weights sharing, and CADN~\cite{long2018conditionalCDAN} improves the discriminator by introducing the condition of discriminative information conveyed in the classifier prediction. Symnets~\cite{zhang2019domainSymnets} builds symmetric classifiers for source and target domains, on which the two-level domain confusion training is based. 
MetaAlign~\cite{wei2021metaalign} brings the MAML~\cite{finn2017modelMAML} to eliminate the conflicts between feature alignment and task objectives. DALN~\cite{chen2022reusingDALN} designs a discriminator-free adversarial paradigm by reusing the category classifier as a discriminator. 

Another popular line on UDA is based on self-training~\cite{saito2017asymmetric,french2017self,zou2018unsupervisedCBST, xie2018learning, zou2019confidenceCRST, gu2020spherical, mei2020instance, liu2021cycle, na2021fixbi, zhang2022udaCAUDA, zhang2022low}, which aims to capture the discriminative information and data structure of the target domain with their pseudo labels. However, the classification models are inevitably susceptible to the adverse effects of noisy inaccurate pseudo-labels. Most studies~\cite{zou2018unsupervisedCBST, zou2019confidenceCRST, liu2021cycle, gu2020spherical, zhang2022udaCAUDA} tend to solve the issue by increasing the reliability of pseudo labels. CRST~\cite{zou2019confidenceCRST} proposes a  regularized self-training paradigm, which composes of label regularization and model regularization. CST~\cite{liu2021cycle} learns to boost the generalization ability of pseudo-labels across domains with a cycle self-training. CA-UDA~\cite{zhang2022udaCAUDA} designs the optimal assignment and pseudo-label refinement to produce reliable pseudo-labels for the target domain by feature clustering and matching. Another category  attempts to leverage the low-confident target samples~\cite{zhang2022low,na2021fixbi}, which also performs excellent domain adaptability.  
Although the above methods have greatly accelerated the development of UDA, they are commonly validated based on convolutional networks.  In this paper, we focus on investigating and improving the domain adaptability of the recent popular vision transformer architecture under the severe domain shift.   



\subsection{Transformer-based UDA}
Transformer architectures~\cite{
carion2020endDETR,
zhu2020deformable,
zheng2021rethinking,
cheng2022masked,
li2022exploring} have achieved remarkable performance on classification, segmentation and object detection, etc, compared with their CNN counterparts. As the pioneering work, Dosovitskiy \etal ~\cite{dosovitskiy2020imageViT} firstly introduce a convolution-free backbone \ieno, ViT, based on pure self-attention. Then, Touvron \etal \cite{touvron2021trainingDEiT} propose a knowledge distillation method through attention to improve the training strategy of ViT. 
Meanwhile, various variants~\cite{liu2021Swin, wang2021pyramidPVT, chu2021twinsTwins} of the transformer are devoted to exploring the optimal architectures for different vision tasks, such as Swin~\cite{liu2021Swin}, Twins~\cite{chu2021twinsTwins}, PVT~\cite{wang2021pyramidPVT}, etc.  

Recently, several works~\cite{xu2021cdtransCDTrans, ma2021exploitingWinTR, yang2021tvtTVT, sun2022safeSSRT, hoyer2022daformerDAformer} have taken a step forward to explore the generalization ability of transformer in UDA. CDTrans~\cite{xu2021cdtransCDTrans} exploits self-attention and cross-attention to explicitly enforce the weight-sharing triple branch transformer to learn domain-specific and -invariant knowledge, and BCAT~\cite{wang2022domainBCAT} extends it to a quadruple-branch transformer. TVT~\cite{yang2021tvtTVT} increases the transferability of  the attention module 
by injecting the knowledge from the discriminator. SSRT~\cite{sun2022safeSSRT} introduces the target domain perturbation and self-refinement, which achieves superior performance in UDA. Differently, 
We aim to improve the domain adaptability of the vision transformer by improving
the richness and flexibility of alignment space from the message-passing perspective, which can be integrated into other existing methods easily.

\begin{figure*}[htp]
    \centering
    \includegraphics[width=0.85\linewidth]{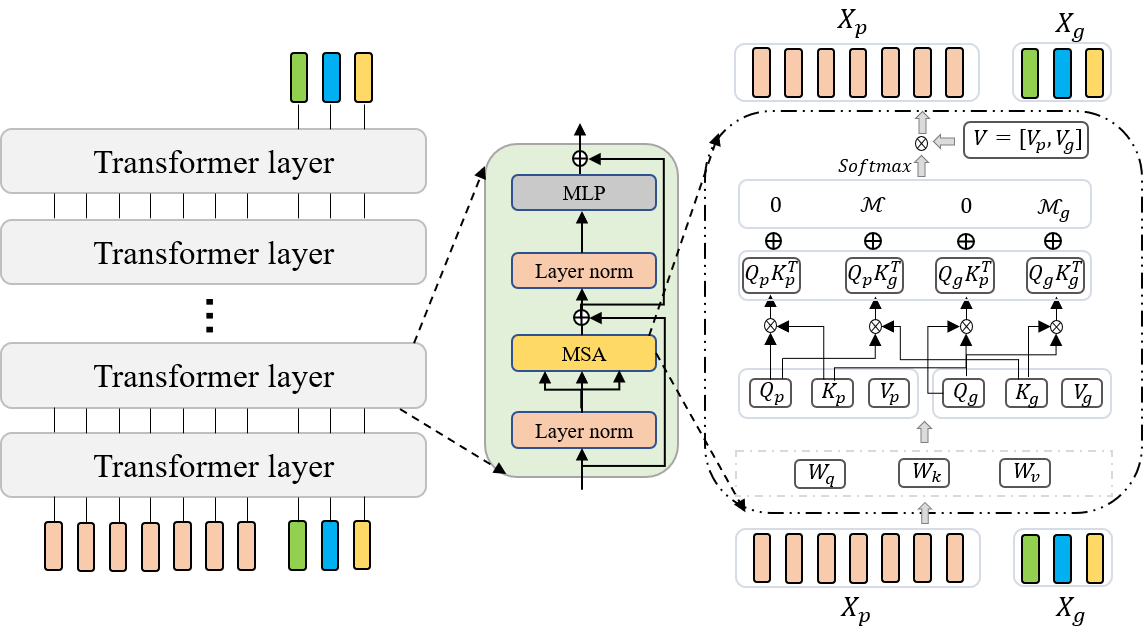}
    \caption{The framework of ViT by incorporating our proposed Semantic-aware Message Broadcasting (SAMB). $X_g$ and $X_p$ are the group tokens and image tokens, respectively. $Q$, $K$, $V$ represent the query, key, and value projected with weights $W_q$, $W_k$, $W_v$.  We implement SAMB with our carefully designed attention mask $\mathcal{M}$, $\mathcal{M}_g$. Here, $\mathcal{M}$ is adaptively generated with the Gumbel softmax~\cite{jang2016categoricalGumbelsoftmax} based on the correlation between the semantics/contents of image tokens and group tokens. We adopt a simple attention layer to obtain the attention matrix $A\in \mathbb{R^{N}}$ to fuse the $N$ group tokens to a token and conduct the classification and domain alignment on this token.}
    \label{fig:framework}
\end{figure*}

\section{Method}
\label{sec:methods}
\tco{In this section, we} first describe the preliminaries for unsupervised domain adaptation and vision transformer. Then, we will introduce our proposed Semantic-Aware Message Broadcasting, and our  training mechanism, respectively. The whole framework we used for transformer-based UDA is shown in Fig.~\ref{fig:framework}, which introduces the learnable group tokens and incorporates the Semantic-aware Message Broadcasting (SAMB) in the multi-head self-attention module. For the output $X_g$ of our framework, we adopt a simple attention layer to obtain the attention matrix $A\in \mathbb{R}^N$ to adaptively fuse the $N$ group tokens to one token for classification and domain alignment. 

\subsection{Preliminaries}
\subsubsection{Unsupervised domain adaptation}
Unsupervised domain adaptation for classification aims to transfer the knowledge learned from labeled source domain $\mathcal{D}_s=\{x^s, y^s\}$ to unlabeled target domain $\mathcal{D}_t=\{x^t\}$, \tco{where $x$ denotes the input image with label $y$}. Feature  alignment~\cite{long2018conditionalCDAN,ganin2016domain_dann,borgwardt2006integratingMMD,chen2019progressive} and self-training~\cite{xu2021cdtransCDTrans,gu2020spherical,mei2020instance,liu2021cycle} are two prominent and popular research directions for UDA. Recently, adversarial learning-based feature alignment~\cite{ganin2016domain_dann} has performed great capability in domain-invariant learning. Particularly, a discriminator $D$ is exploited to distinguish the source and target domain. To fool the discriminator, the feature extractor $G$ is enforced to learn the domain-invariant representation. The objectives for adversarial feature alignment-based UDA is as~\cite{ganin2016domain_dann}:
\begin{equation}
    \label{eq:ada_uda}
     min ~ \mathcal{L}_{cls}(C(G(x^s)), y^s) - \mathcal{L}_D(x_s, x_t),
\end{equation}
where $\mathcal{L}_{cls}$ and $C$ are the supervised classification loss and a classifier, respectively. The domain classification loss $\mathcal{L}_D$ is typically defined as:

\begin{equation}
     \label{eq:l_d}
\begin{split}
    \mathcal{L}_D(x_s, x_t) = & -\mathbb{E}_{x_s \sim D_s}[log(D(G(x_s)))]  \\
    & -\mathbb{E}_{x_t \sim D_t}[log(1-D(G(x_t)))].
\end{split}
\end{equation}
\tcb{The domain classifier is trained to minimize $\mathcal{L}_D$.}

Another popular line for UDA is exploiting \tcb{pseudo-label based} self-training, where the pseudo-labels of the target domain are used for learning the discriminative information and data structure. 
In this paper, we obtain the pseudo-labels of the target domain with the method from~\cite{xu2021cdtransCDTrans}. 
\begin{table*}[htp]
\centering
\caption{Classification accuracy (\%) of different UDA methods on VisDA-2017~\cite{peng2017visdaViSDA2017}.
}
\setlength{\tabcolsep}{2mm}{\begin{tabular}{cccccccccccccc}
\hline
Method                              & plane                & bcycl                                 & bus                  & car                                   & horse                                 & knife                                 & mcycl                & person                                & plant                                 & sktbrd                                & train                                & truck                                 & Avg.                                  \\ \hline
ResNet-101~\cite{he2016deepResNet}                          & 55.1                 & 53.3                                  & 61.9                 & 59.1                                  & 80.6                                  & 17.9                                  & 79.7                 & 31.2                                  & 81.0                                  & 26.5                                  & 73.5                                 & 8.5                                   & 52.4                                  \\
DANN~\cite{ganin2016domain_dann}                                & 81.9                 & 77.7                                  & 82.8                 & 44.3                                  & 81.2                                  & 29.5                                  & 65.1                 & 28.6                                  & 51.9                                  & 54.6                                  & 82.8                                 & 7.8                                   & 57.4                                  \\
CDAN~\cite{long2018conditionalCDAN}                                & 85.2                 & 66.9                                  & 83.0                 & 50.8                                  & 84.2                                  & 74.9                                  & 88.1                 & 74.5                                  & 83.4                                  & 76.0                                  & 81.9                                 & 38.0                                  & 73.9                                  \\
SAFN~\cite{xu2019largerSAFN}                                & 93.6                 & 61.3                                  & 84.1                 & 70.6                                  & 94.1                                  & 79.0                                  & 91.8                 & 79.6                                  & 89.9                                  & 55.6                                  & 89.0                                 & 24.4                                  & 76.1                                  \\
SWD~\cite{lee2019slicedSWD}                                 & 90.8                 & 82.5                                  & 81.7                 & 70.5                                  & 91.7                                  & 69.5                                  & 86.3                 & 77.5                                  & 87.4                                  & 63.6                                  & 85.6                                 & 29.2                                  & 76.4                                  \\
SHOT~\cite{liang2020weSHOT}                                & 94.3                 & 88.5                                  & 80.1                 & 57.3                                  & 93.1                                  & 94.9                                  & 80.7                 & 80.3                                  & 91.5                                  & 89.1                                  & 86.3                                 & 58.2                                  & 82.9                                  \\ \hline
CDTrans-B~\cite{xu2021cdtransCDTrans}                           & 97.1                 & 90.5                                  & 82.4                 & 77.5                                  & 96.6                                  & 96.1                                  & 93.6                 & \textcolor{red}{\textbf{88.6}}                                  & 97.9                                  & 86.9                                  & 90.3                                 & \textcolor{red}{\textbf{62.8}}                                  & 88.4                                  \\
BCAT-DTF~\cite{wang2022domainBCAT}                            & \textcolor{red}{\textbf{99.1}}                 & 91.6                                  & 86.6                 & 72.3                                  & 98.7                                  & 97.9                                  & \textcolor{red}{\textbf{96.5}}                 & 82.3                                  & 94.2                                  & 96.0                                  & 93.9                                 & 61.3                                  & 89.2                                  \\
WinTR-B~\cite{ma2021exploitingWinTR}                             & 98.7                 & 91.2                                  & \textcolor{red}{\textbf{93.0}}                 & \textcolor{red}{\textbf{91.9}}                                  & 98.1                                  & 96.1                                  & 94.0                 & 72.7                                  & 97.0                                  & 95.5                                  & 95.3                                 & 57.9                                  & 90.1                                  \\
TVT~\cite{yang2021tvtTVT}                                 & 92.92                & 85.58                                 & 77.51                & 60.48                                 & 93.60                                 & 98.17                                 & 89.35                & 76.40                                 & 93.56                                 & 92.02                                 & 91.69                                & 55.73                                 & 83.92                                 \\ \hline \hline
Baseline-S                          & \multicolumn{1}{c}{97.75} & \multicolumn{1}{c}{68.81}                  & \multicolumn{1}{c}{82.99} & \multicolumn{1}{c}{68.28}                  & \multicolumn{1}{c}{95.01}                  & \multicolumn{1}{c}{96.92}                  & \multicolumn{1}{c}{95.89} & \multicolumn{1}{c}{73.57}                  & \multicolumn{1}{c}{86.66}                  & \multicolumn{1}{c}{82.03}                  & \multicolumn{1}{c}{94.22}                 & \multicolumn{1}{c}{31.04}                  & \multicolumn{1}{c}{81.10}                  \\
\rowcolor[gray]{0.9} \multicolumn{1}{c}{+Ours} & \multicolumn{1}{c}{98.22$_\uparrow$} & \multicolumn{1}{c}{77.50$_\uparrow$}                  & \multicolumn{1}{c}{84.54$_\uparrow$} & \multicolumn{1}{c}{65.60}                  & \multicolumn{1}{c}{96.95$_\uparrow$}                  & \multicolumn{1}{c}{96.82}                  & \multicolumn{1}{c}{95.63} & \multicolumn{1}{c}{80.55$_\uparrow$}                  & \multicolumn{1}{c}{93.67$_\uparrow$}                  & \multicolumn{1}{c}{90.09$_\uparrow$}                  & \multicolumn{1}{c}{93.22}                 & \multicolumn{1}{c}{47.57$_\uparrow$}                  & \multicolumn{1}{c}{85.03$_\uparrow$}                  \\ \hline

Baseline-B                          & \multicolumn{1}{c}{98.55} & \multicolumn{1}{c}{82.59}                  & \multicolumn{1}{c}{85.97} & \multicolumn{1}{c}{57.07}                  & \multicolumn{1}{c}{94.93}                  & \multicolumn{1}{c}{97.20}                  & \multicolumn{1}{l}{94.58} & \multicolumn{1}{c}{76.68}                  & \multicolumn{1}{c}{92.11}                  & \multicolumn{1}{c}{96.54}                  & \multicolumn{1}{c}{94.31}                 & \multicolumn{1}{c}{52.24}                  & \multicolumn{1}{c}{85.23}                  \\

\rowcolor[gray]{0.9} \multicolumn{1}{c}{+Ours} & \multicolumn{1}{c}{98.96$_\uparrow$} & \multicolumn{1}{c}{85.50$_\uparrow$}                  & \multicolumn{1}{c}{84.09} & \multicolumn{1}{c}{67.30$_\uparrow$}                  & \multicolumn{1}{c}{97.78$_\uparrow$}                  & \multicolumn{1}{c}{97.40$_\uparrow$}                  & \multicolumn{1}{c}{94.20} & \multicolumn{1}{c}{83.50$_\uparrow$}                  & \multicolumn{1}{c}{95.34$_\uparrow$}                  & \multicolumn{1}{c}{94.74}                  & \multicolumn{1}{c}{93.20}                 & \multicolumn{1}{c}{55.97$_\uparrow$}                  & \multicolumn{1}{c}{87.33$_\uparrow$}                  \\ \hline



SSRT-B~\cite{sun2022safeSSRT}                              & 98.93                & 87.60                                 & 89.10                & 84.77                                 & 98.34                                 & 98.70                                 & 96.27                & 81.08                                 & 94.86                                 & 97.90                                 & 94.50                                & 43.13                                 & 88.76                                 \\
\rowcolor[gray]{0.9} \multicolumn{1}{c}{+Ours}     & 98.85       &  \textcolor{red}{\textbf{92.12}}$_\uparrow$ & 87.40        &  87.87$_\uparrow$ &  \textcolor{red}{\textbf{98.83}}$_\uparrow$ &  \textcolor{red}{\textbf{98.94}}$_\uparrow$ & 95.74       & 85.23$_\uparrow$ &  \textcolor{red}{\textbf{97.98}}$_\uparrow$ &  \textcolor{red}{\textbf{98.68}}$_\uparrow$ &   \textcolor{red}{\textbf{95.30}}$_\uparrow$ & 47.93$_\uparrow$ & \textcolor{red}{\textbf{90.41}}$_\uparrow$ \\ \hline
\end{tabular}}
\label{tab:visda-2017}
\end{table*}
\subsubsection{Vision Transformer}
Vision Transformer has been broadly exploited in abundant vision tasks due to its great power of inductive bias and flexibility. For a given image, ViT divides it into a sequence of image patches $X_p=[X_p^1, X_p^2, ..., X_p^M]$, where $M$ is the number of patches. A learnable classification token $X_{cls}$ is appended as the task prior~\cite{dosovitskiy2020imageViT} to assist the classification. Self-attention~\cite{vaswani2017attentionisallyouneed}, as the most important structure in ViT, aims to leverage the dependency between different tokens to aggregate the valuable information for each token from other tokens as follows:

\begin{equation}
    \label{eq:selfattn}
    Attn (Q, K, V) = softmax(\frac{QK^T}{\sqrt{d_k}})V,
\end{equation}
where query, key and value (\ieno, $Q$, $K$, $V$) are obtained by projecting the input tokens $X=[X_{cls}, X_p]$ with three learnable projectors $W_Q$, $W_K$, $W_V$.

\subsection{Dynamic Semantic-aware Message Broadcasting}
For the unsupervised domain adaptation (UDA) of the vision transformer, the domain alignment is generally conducted in the feature space used for classification (\ieno, the class token). As shown in Fig.~\ref{fig:concept_comparion} (a), there are two typical message-passing processes for the class token, \ieno, message aggregation, and message broadcasting, respectively. Concretely, \textit{the message aggregation process denotes the class token aggregates the global information from all image tokens, and the message broadcasting process represents that the class token broadcasts its knowledge to each local image token.} 
These processes enable the class token to learn global contextual classification \tcb{information but fail to identify and exploit the rich/diverse semantics of different spatial regions. Alignment only on this class token restricts the effectiveness of domain adaptation.} 
In this paper, we aim to improve the richness and flexibility of the alignment features from the message-broadcasting perspective. 

\subsubsection{Semantic-aware Message Broadcasting} 
\label{sec:samb}
To encourage the features for alignment to capture more informative and diverse information, we propose Semantic-aware Message Broadcasting for Transformer-based Unsupervised Domain Adaptation (UDA). It is noteworthy that the semantics of images are always diverse \tcb{for different spatial regions}.
For instance, the objects and background are \tcb{located} in different regions. \tcb{How to enable efficient information interaction and explore the rich semantics for alignment is still under-explored.}  
In this work, as shown in Fig.~\ref{fig:concept_comparion}(b), we set a group of learnable group tokens $\{X_g^i\}_{i=1}^N$ as semantic-aware nodes. Each \emph{group token} (node) is responsible for the message-passing route of \tcb{regions of some specific semantics},
which encourages \tcb{the network} to learn semantic-aware informative and diverse knowledge for domain alignment. 

Particularly, we reform the input of the attention module as $X^\dagger = [X_g^1, ..., X_g^N, X_p^1, X_p^2, ..., X_p^M]$, where $N$ and $M$ denote the number of \tcb{learnable} group tokens and image tokens\tcb{, where $N < M$}.
\tcb{Instead of connecting to all image tokens, a group token gathers information from all image tokens but broadcasts to only a set of related image tokens.} 
We achieve the assignment of the group tokens (nodes) for the semantic-aware message broadcasting in the attention module \tcb{through} the attention mask $\mathcal{M}$. It is noteworthy that the message passing in the original attention module is established through the correlation matrix $QK^T$ in Eq.~\ref{eq:selfattn}, where the value $q_ik_j$ in $QK^T$ describes how many messages from the $j^{th}$ token are passed to the $i^{th}$ token. That means when we mask the $q_ik_j$ by adding the value $-\infty$ to it, the message passing path from $j^{th}$ token to $i^{th}$ token will be cut off since the value becomes zero after $softmax$ operation in Eq.~\ref{eq:selfattn}. 

Now, let us identify the components in the correlation matrix $QK^T$ that are responsible for the message broadcasting and aggregation of the group tokens. Here, we define the $\{Q_g, K_g\}$,  $\{Q_p, K_p\}$ as the pairs of key and value projected with region prior $X_g=[X_g^1, ..., X_g^N]$ and image tokens $X_p$, respectively. And thus, we can decouple the correlation metric $QK^T$ into four components:
\begin{equation}
    QK^T= \begin{bmatrix}
 Q_g K_g^T & Q_g K_p^T \\
  Q_p K_p^T & Q_p K_g^T,
  \end{bmatrix}.
  \label{equ:corrematrix}
\end{equation}
where $Q_pK_g^T$, and $Q_gK_p^T$ are responsible for the message broadcasting, and the message aggregation for the group tokens, respectively. Based on the above analysis, we can easily achieve the semantic-aware message broadcasting by designing a proper attention mask $\mathcal{M}$ for $Q_pK_g^T$. 

An intuitive strategy for this is to design a hand-craft mask $\mathcal{M}$ for $Q_pK_g^T$ based on different regions since the contents/semantics of different regions are commonly diverse. Concretely, we can divide one image into $N$ regions, and each group token is responsible for the message broadcasting of one region. For example, given $N=2$ learned group tokens and $M=4$ image tokens, the attention mask $\mathcal{M}$ can be set as:
\begin{equation}
     \mathcal{M} = \begin{bmatrix}
      0 & 0 & -\infty & -\infty  \\
      -\infty & -\infty & 0 & 0
  \end{bmatrix}^T_{N\times M}
  \label{eq:naive_mask}
\end{equation}
 However, it is not optimal for the hand-crafted assignment, which lacks enough flexibility and dynamics. We aim to achieve the dynamic assignment for group tokens based on the characteristics of different image tokens, which is achieved with our proposed dynamic assignment in Sec.~\ref{sec:da} 
 

Furthermore, to avoid the intervention of different group tokens and retain the message scale in the attention module, we also cut off the message passing paths of different group tokens by adding them an attention mask $\mathcal{M}_g$ as: 
\begin{equation}
    \mathcal{M}_g = \begin{bmatrix}
 0 & & -\infty \\
  & \ddots &  \\
  -\infty & & 0
  \end{bmatrix}_{N\times N}.
\end{equation}
Taking the above schemes, we can obtain the attention matrix for semantic-aware adaptive message broadcasting:
\begin{equation}
\begin{aligned}
    & Attn(Q,K,V) = \\ & softmax(\begin{bmatrix}
 Q_g K_g^T+\mathcal{M}_g & Q_g K_p^T \\
  Q_p K_p^T & Q_p K_g^T + \mathcal{M}
  \end{bmatrix}/{\sqrt{d_k}})V
\end{aligned}
\end{equation}

\subsubsection{Dynamical Assignment} 
\label{sec:da}
As described in the above Sec.~\ref{sec:samb}, the semantic-aware message broadcasting is achieved with the attention mask $\mathcal{M}$. To achieve the dynamic assignment of group tokens, it is necessary to adaptively generate the discrete mask $\mathcal{M}$ based on the contents of images.
\begin{figure*}[htp]
    \centering
    \includegraphics[width=0.8\linewidth]{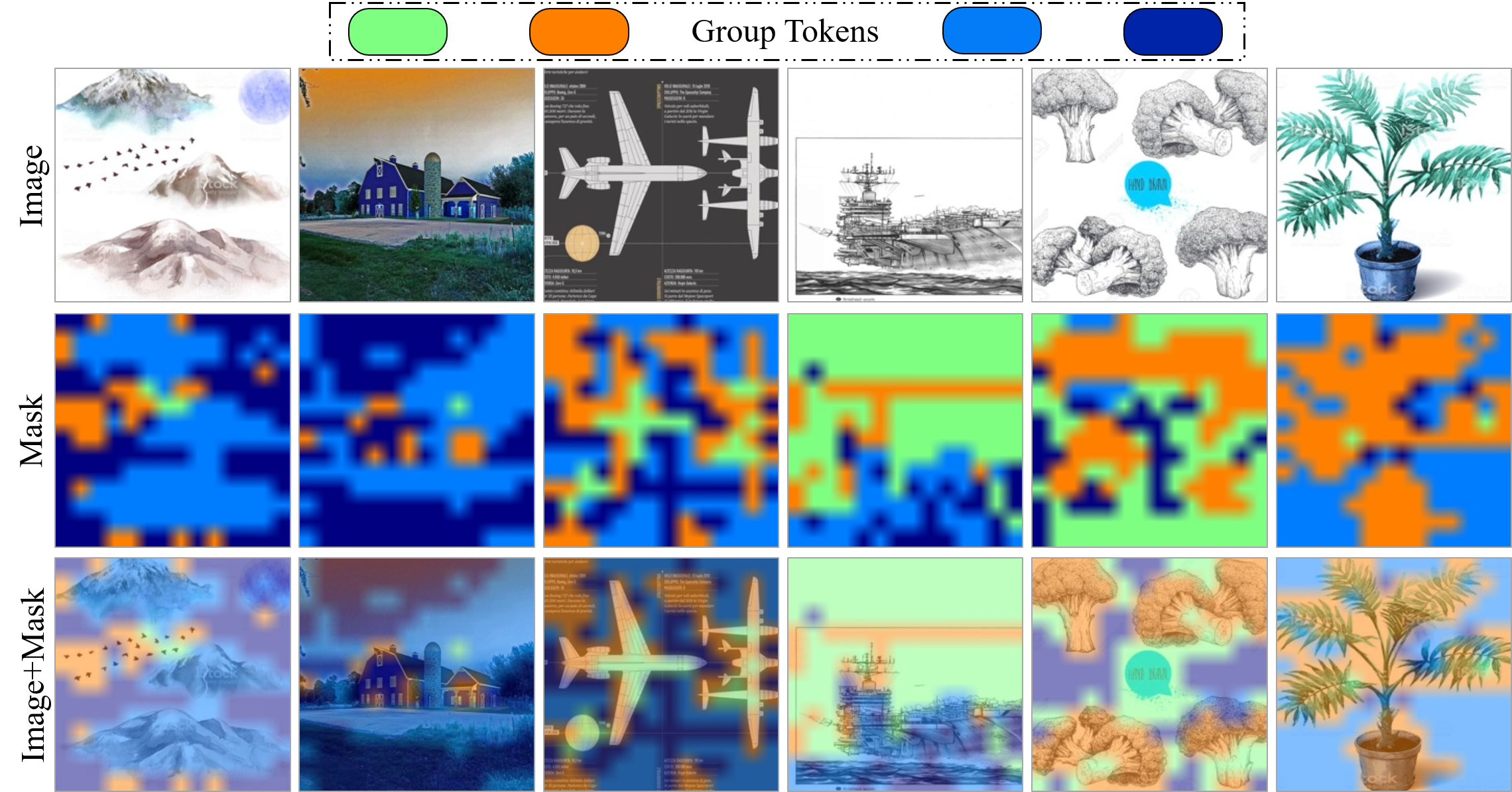}
    \caption{Visualization of the assignment for group tokens in each image. The first row shows the original image, and the second row represents the assignment map of the different group tokens with different colors. The third row is the superposition of the top two rows.}  
    \label{fig:region_assignment}
\end{figure*}
Particularly, for each image region, it requires to be assigned a specific and optimal group token. Therefore, each line $m_i^*$ of $\mathcal{M}\in \mathbb{R}^{M\times N}$  must be a one-hot vector with dimension $N$, where the gradients cannot be directly backward through the discrete values. Inspired by~\cite{xu2022groupvit, van2017neuralVQVAE, jang2016categoricalGumbelsoftmax}, we enable this process differential with one reparametrization trick, Straight-through Gumbel-Softmax~\cite{jang2016categoricalGumbelsoftmax}, which can be represented as:
\begin{equation}
\begin{split}
      m_{ij} &= \frac{exp((Q_p^i {K_g^j}^T + u_{ij})/\tau)}{\sum_{j=1}^N exp((Q_p^i {K_g^j}^T+u_{ij})/\tau)}, \\ 
       &~ u_{ij} \sim Gumbel(0, 1), ~ 0\le i \le M, ~ 0 \le j \le N. 
\end{split}
\end{equation}
Then we can obtain each line $m_i^*$ of dynamic mask $\mathcal{M}$ with hard attention as:
\begin{equation}
\begin{split}
     m_i &= \mathrm{one}\mbox{-}\mathrm{hot}(m_i) + m_i - sg(m_i) \\
     m_i^* &= -\infty * (1-m_i),
\end{split}
\end{equation}
where $sg$ represents the stop gradient operation. Based on this strategy, we can achieve the purpose of dynamic assignment for group tokens with $\mathcal{M}_3=\{m_i\}_{i=1}^{M}$. 

\subsubsection{Discussion on Message Scale}
\label{sec:ms}
Another advantage of our proposed semantic-aware message broadcasting is the message scale in the original ViT remains unchanged. We have an interesting and critical observation that a changed message scale is harmful to UDA. The reason we guess is that it impairs the knowledge learned from pre-trained ImageNet. In this paper, the message scale refers to the number of $V$ used to refine each token in the attention module. For original ViT~\cite{dosovitskiy2020imageViT}, each token receives the information from $M+1$ tokens, and the message scale is $M+1$. In our method, each region token $X_g^i$ only aggregates the information from itself and other image tokens $X_p$ since attention mask, where the message scale is as $M+1$. Besides, the message scale received by each image token $X_p^j$ is also $M+1$, which is composed of one corresponding group token and $M$ image tokens. 

We also investigate different message-passing strategies for Transformer-based UDA, \ieno, the semantic-aware message aggregation for $Q_gK_p^T$ in Eq.~\ref{equ:corrematrix} (termed as``SAMG"), global and local message aggregation and broadcasting (denoted as ``G-L"), and global message aggregation and broadcasting (denoted as ``G-G"). However, they cannot achieve a satisfying performance since they cause the modification of the message scale. We will clarify it in the ablation studies of experiments in ~\ref{sec:ab_ms}.  
 \begin{algorithm}[t]
\caption{Training Mechanism for Transformer-based UDA}
\label{alg: training}
\begin{algorithmic}[1]
\State \textbf{Input:} Source and target data $\mathcal{D}_s$ and $\mathcal{D}_t$ 
\State \textbf{Init:} learning rate: $\alpha$, parameters $\Psi=\{ \theta_G,\theta_D,\theta_C\}$
\For {t \textbf{in} iterations$_1$}
\State Compute the domain alignment loss $\mathcal{L}_D$ and supervised loss $\mathcal{L}_{cls}$.
\State  Update $\theta_G$, $\theta_C$ w.r.t. $\mathcal{L}_D$ and $\mathcal{L}_{cls}$: \\
    \quad \quad \quad         $\Psi^{t+1} \xleftarrow{} \Psi^t - \alpha \nabla_{\Psi^t}(\mathcal{L}_{cls}+\mathcal{L}_D)$
\EndFor 
\For {t \textbf{in} iterations$_2$}
\State Obtaining the pseudo-labels $y_t^*$ 
\State 
Compute the domain alignment loss $\mathcal{L}_D$ and supervised loss $\mathcal{L}_{cls}$ on $\{x_s, y_s\}$ and $\{x_t, y_t^*\}$
\State  Update $\theta_G$, $\theta_C$ w.r.t. $\mathcal{L}_D$ and $\mathcal{L}_{cls}$: \\
    \quad \quad \quad         $\Psi^{t+1} \xleftarrow{} \Psi^t - \alpha \nabla_{\Psi^t}(\mathcal{L}_{cls}+\mathcal{L}_D)$
\EndFor 
\end{algorithmic}
\end{algorithm}

\subsection{Training Mechanism}
Existing works on Transformer-based UDA have revealed the  priority of Adversarial-based feature Alignment (ADA) and Pseudo-label based Self-training (PST). However, few works take further exploration for the cooperation of ADA and PST on Transformer-based UDA. In this paper, we aim to find an easy but effective cooperation strategy for ADA and PST. To achieve this goal, we investigate different combinations of ADA and PST with the following schemes: 1) only ADA or PST, 2) (ADA, PST) (\textit{i.e.}, train the network with ADA and PST jointly, 3) ADA $\xrightarrow{}$ PST \tcb{(\textit{i.e.}, train the network with ADA first and then with PST)}, 4) PST $\xrightarrow[]{}$ ADA, 5) ADA $\xrightarrow[]{}$ (ADA, PST). The specific methods we used for ADA and PST in our paper are from DANN~\cite{ganin2016domain_dann, sun2022safeSSRT}, and CDTrans~\cite{xu2021cdtransCDTrans}. For ADA, one gradient reversal layer (GRL)~\cite{ganin2016domain_dann} is leveraged to connect the gradients between $G$ and $D$ via multiplying the gradient from $D$ by a negative constant. For PST, instead of collecting the pseudo-labels with one-hot labeling from the pre-trained model, which is noisy and unreliable, we collect the pseudo-labels based on the weighted K-means feature clustering as:
\begin{equation}
    c_k = \frac{\sum_{i\in T}\delta_k^i G(x_t^i)}{\sum_{i \in T}\delta_k^i}, y_t^i = argmin_k d(c_k, G(x_t^i)),
\end{equation}
where $c_k$ denotes the weighted cluster center with class $k$, and $\delta_k^i$ represents the probability of $i^{th}$ sample $x_t^i$ in the $k^{th}$ class. $d$ refers to the distance between two features. After obtaining the preliminary pseudo-labels $y_t$, more accurate cluster center $c_k^*$ and pseudo labels $y_t^*$ can be computed as:
\begin{equation}
    c_k^* = \frac{\sum_{i\in T} {\mathbbm{1}} (y_t^i=k) G(x_t^i)}{\sum_{i\in T} \mathbbm{1} (y_t^i=k) }, y_t^* = argmin_k d(c_k^*, G(x_t)).
\end{equation}
\tcb{We have studied the efficiency of different schemes and show the results in Fig.~\ref{fig:curve_training_mechanism}.} Based on the observations, we have the following valuable conclusions. 1) It is easy to suffer from model collapse by optimizing the network with PST and ADA together in a simplified manner. The reason for that should be the strong capability for inductive bias towards the target distribution. However, the pseudo-labels produced by the network are unreliable in this process.
2) More reliable pseudo labels (\ieno, more strong initial models for target domain) result in a better generalization capability for PST.
3) ADA is complementary to PST. It can achieve the best performance by training the model with ADA first, and then optimizing the model with ADA and PST together. 
In this paper, we utilize the ADA $\xrightarrow[]{}$ (ADA,PST) as our training mechanism. The details of our training mechanism are described in algorithm~\ref{alg: training}.

\begin{table*}[htp]
\centering
\caption{Classification accuracy (\%) of different UDAs on Office-Home~\cite{venkateswara2017deepOffice-Home}.}
\setlength{\tabcolsep}{1mm}{\begin{tabular}{cccccccccccccc}
\hline
Method                              & Ar$\xrightarrow{}$Cl                & Ar$\xrightarrow[]{}$Pr                                 & Ar$\xrightarrow[]{}$Rw   & Cl$\xrightarrow[]{}$Ar                                  & Cl$\xrightarrow[]{}$Pr   & Cl$\xrightarrow[]{}$Rw    & Pr$\xrightarrow[]{}$Ar  & Pr$\xrightarrow[]{}$Cl    &  Pr$\xrightarrow[]{}$Rw     & Rw$\xrightarrow[]{}$Ar  & Rw$\xrightarrow[]{}$Cl  & Rw$\xrightarrow[]{}$Pr  & Avg.                                  \\ \hline
ResNet-50~\cite{he2016deepResNet} & 34.9 & 50.0 & 58.0 & 37.4 & 41.9 & 46.2 & 38.5 & 31.2 & 60.4 & 53.9 & 41.2 & 59.9 & 41.6 \\
CDAN+E~\cite{long2018conditionalCDAN} & 50.7 & 70.6 & 76.0 & 57.6 & 70.0 & 70.0 & 57.4 & 50.9 & 77.3 & 70.9 & 56.7 & 81.6 & 65.8\\
SAFN~\cite{xu2019largerSAFN} & 52.0 & 71.7  & 76.3 & 64.2 & 69.9 & 71.9 & 63.7 & 51.4  & 77.1  & 70.9 & 57.1 & 81.5 & 67.3  \\
CDAN+TN~\cite{wang2019transferableCDAN+TN} & 50.2 & 71.4 & 77.4  & 59.3  & 72.7  & 73.1 & 61.0  & 53.1 & 79.5 & 71.9  & 59.0  & 82.9 & 67.6\\
SHOT~\cite{liang2020weSHOT}  & 57.1 & 78.1 & 81.5  & 68.0  & 78.2 & 78.1 & 67.4 & 54.9  & 82.2 & 73.3 & 58.8 & 84.3 & 71.8\\
HDA+ToAlign~\cite{wei2021toalign} & 57.9 & 76.9 & 80.8 & 66.7 & 75.6 & 77.0 & 67.8 & 57.0 & 82.5 & 75.1 & 60.0 & 84.9 & 72.0 \\
DCAN+SCDA~\cite{li2021semanticDCAN+SCDA} & 60.7 & 76.4 & 82.8 & 69.8 & 77.5  & 78.4 & 68.9 & 59.0 & 82.7 & 74.9 & 61.8 & 84.5  & 73.1 \\ 
TransPar-MCC~\cite{han2022learning} & 58.9 & 80.7 & 83.4 & 67.6 & 77.6 & 78.6 & 68.2 & 55.7 & 82.3 & 75.1 & 62.5 & 86.2 & 73.1 \\ \hline
CDTrans-B~\cite{xu2021cdtransCDTrans}   & 68.8 & 85.0 & 86.9 & 81.5 & 87.1 & 87.3 & 79.6  & 63.3 & 88.2 & 82.0 & 66.0 & 90.6 & 80.5 \\
WinTR-S~\cite{ma2021exploitingWinTR} & 65.3 & 84.1 & 85.0 & 76.8 & 84.5 & 84.4 & 73.4 & 60.0 & 85.7 & 77.2 & 63.1 & 86.8 & 77.2 \\
TVT~\cite{yang2021tvtTVT}  & 74.89 & 86.82 & 89.47 & 82.78 & 87.95 & 88.27 & 79.81 & 71.94 & 90.13 & 85.46 & 74.62 & 90.56 & 83.56  \\ \hline \hline
Baseline-S                          & \multicolumn{1}{c}{62.88} & \multicolumn{1}{c}{79.52}                  & \multicolumn{1}{c}{86.09} & \multicolumn{1}{c}{77.62}                  & \multicolumn{1}{c}{81.91}                  & \multicolumn{1}{c}{82.96}                  & \multicolumn{1}{c}{75.40} & \multicolumn{1}{c}{63.13}                  & \multicolumn{1}{c}{87.53}                  & \multicolumn{1}{c}{80.63}                  & \multicolumn{1}{c}{65.22}                 & \multicolumn{1}{c}{87.11}                  & \multicolumn{1}{c}{77.50}                  \\
\rowcolor[gray]{0.9}\multicolumn{1}{c}{+Ours} & \multicolumn{1}{c}{65.66$_\uparrow$} & \multicolumn{1}{c}{84.01$_\uparrow$}                  & \multicolumn{1}{c}{87.74$_\uparrow$} & \multicolumn{1}{c}{80.22$_\uparrow$}                  & \multicolumn{1}{c}{84.14$_\uparrow$}                  & \multicolumn{1}{c}{86.73$_\uparrow$}                  & \multicolumn{1}{c}{78.78$_\uparrow$} & \multicolumn{1}{c}{64.40$_\uparrow$}                  & \multicolumn{1}{c}{88.57$_\uparrow$}                  & \multicolumn{1}{c}{82.41$_\uparrow$}                  & \multicolumn{1}{c}{67.03$_\uparrow$}                 & \multicolumn{1}{c}{87.99$_\uparrow$}                  & \multicolumn{1}{c}{79.81$_\uparrow$}                  \\ \hline
Baseline-B                          & \multicolumn{1}{c}{66.96} & \multicolumn{1}{c}{85.74}                  & \multicolumn{1}{c}{88.07} & \multicolumn{1}{c}{80.06}                  & \multicolumn{1}{c}{84.12}                  & \multicolumn{1}{c}{86.67}                  & \multicolumn{1}{c}{79.52} & \multicolumn{1}{c}{67.03}                  & \multicolumn{1}{c}{89.44}                  & \multicolumn{1}{c}{83.64}                  & \multicolumn{1}{c}{70.15}                 & \multicolumn{1}{c}{91.17}                  & \multicolumn{1}{c}{81.05}                  \\

\rowcolor[gray]{0.9} \multicolumn{1}{c}{+Ours} & \multicolumn{1}{c}{68.66$_\uparrow$} & \multicolumn{1}{c}{84.97}                  & \multicolumn{1}{c}{88.85$_\uparrow$} & \multicolumn{1}{c}{80.80$_\uparrow$}                  & \multicolumn{1}{c}{85.99$_\uparrow$}                  & \multicolumn{1}{c}{88.30$_\uparrow$}                  & \multicolumn{1}{c}{81.58$_\uparrow$} & \multicolumn{1}{c}{68.71$_\uparrow$}                  & \multicolumn{1}{c}{90.22$_\uparrow$}                  & \multicolumn{1}{c}{84.14$_\uparrow$}                  & \multicolumn{1}{c}{70.88$_\uparrow$}                 & \multicolumn{1}{c}{91.12}                  & \multicolumn{1}{c}{82.02$_\uparrow$}                  \\ \hline
SSRT-S~\cite{sun2022safeSSRT} & 67.03 & 84.21 & 88.32 & 79.85 & 84.28 & 87.58 & 80.72 & 66.03 & 88.27 & 82.04 & 69.44 & 89.86 & 80.64 \\
\rowcolor[gray]{0.9}+Ours & $70.24_\uparrow$ & $85.60_\uparrow$ & $89.49_\uparrow$ & $81.83_\uparrow$ & $87.77_\uparrow$ & $89.05_\uparrow$ & 80.63 & $68.29_\uparrow$ & $89.35_\uparrow$ & $82.37_\uparrow$ & $70.38_\uparrow$ & $90.11_\uparrow$ & $82.09_\uparrow$ \\ \hline
SSRT-B~\cite{sun2022safeSSRT}  & 75.17 & \textcolor{red}{\textbf{88.98}} & \textcolor{red}{\textbf{91.09}} & 85.13 & 88.29 & 89.95 & 85.04 & 74.23 & 91.26 & 85.70 & 78.58 & 91.78 & 85.43 \\
\rowcolor[gray]{0.9} \multicolumn{1}{c}{+Ours}  & $\textcolor{red}{\textbf{78.56}}_\uparrow$ & 88.49 & 90.5 & $\textcolor{red}{\textbf{85.95}}_\uparrow$ & $\textcolor{red}{\textbf{90.02}}_\uparrow$ & $\textcolor{red}{\textbf{90.50}}_\uparrow$ & $\textcolor{red}{\textbf{85.74}}_\uparrow$ & $\textcolor{red}{\textbf{75.67}}_\uparrow$ & $\textcolor{red}{\textbf{91.37}}_\uparrow$ & $\textcolor{red}{\textbf{85.99}}_\uparrow$  & $\textcolor{red}{\textbf{78.74}}_\uparrow$ &  $\textcolor{red}{\textbf{92.84}}_\uparrow$ & $\textcolor{red}{\textbf{86.20}}_\uparrow$\\  \hline
\end{tabular}}
\label{tab:office-home}
\end{table*}

\section{Experiments}
\label{sec:experiments}
In this section, we first describe the datasets and the implementation details for SAMB in Sec.~\ref{sec:data_imp}. Then, we compare our proposed SAMB with the state-of-the-art UDA methods on several commonly-used datasets, including DomainNet~\cite{peng2019momentDomainNet}, OfficeHome~\cite{venkateswara2017deepOffice-Home}, and VisDA-2017 in Sec.~\ref{sec:sota}. In Sec.~\ref{sec:ab}, we conduct ablation studies for different message-passing schemes, the number of group tokens and different training mechanisms. We also validate the effectiveness of our proposed SAMB by visualizing the assignment of the group tokens and the t-SNE visualization of features in Sec.~\ref{sec:vis}. The complexity analysis is conducted in Sec.~\ref{sec:complexity}. \textit{It is noteworthy that the SAMB refers to our Semantic-aware Message Broadcasting with the dynamic assignment (SAMB-D) if not mentioned.} Only in the ablation studies for different message-passing schemes, we distinguish between using dynamic assignment or not with SAMB-D/SAMB.
\subsection{Datasets and Implementation}
\label{sec:data_imp}
\subsubsection{Datasets}
We validate our method on three commonly-used datasets for UDA, \ieno, DomainNet~\cite{peng2019momentDomainNet}, Office-Home~\cite{venkateswara2017deepOffice-Home} and VisDA-2017~\cite{peng2017visdaViSDA2017}. 1) DomainNet, as a large-scale dataset, contains about 600,000 images across 345 categories. There are 6 domains with a large domain gap in DomainNet: Clipart (C), Infograph (I), Painting (P), Quickdraw (Q), Real (R) and Sketch (S). 
Following previous works~\cite{sun2022safeSSRT,xu2021cdtransCDTrans}, 
we evaluate methods on the typical settings, \ieno, one source domain to one target domain. 
2) Office-Home is composed of four domains, including Art (Ar), Clipart (Cl), Product (Pr), and Real-World (Rw), and each domain contains 65 categories in office and home environments. Following the works~\cite{sun2022safeSSRT,wei2021toalign,xu2021cdtransCDTrans}, we experiment with it by setting one domain as the source and another domain as the target for UDA. 3) VisDA-2017~\cite{peng2017visdaViSDA2017} focuses on the simulation-to-reality shift, which covers about 280,000 images across 12 categories. The source images are synthetic, and the target images are from real scenarios.
\begin{table*}[]
\centering
\caption{Classification accuracy (\%) of different UDA methods on DomainNet~\cite{peng2019momentDomainNet}.}
\setlength{\tabcolsep}{0.7mm}{\begin{tabular}{cccccccccccccccccccccccccc}
\hline
\multicolumn{1}{|c|}{\makecell[c]{ResNet-\\101}} & clp                  & inf                      & pnt                  & qdr                  & rel                  & skt                  & \multicolumn{1}{c|}{Avg.} & \multicolumn{1}{c|}{} & \multicolumn{1}{c|}{CDAN}     & clp                  & inf                  & pnt                  & qdr                  & rel                  & skt                  & \multicolumn{1}{c|}{Avg.} & \multicolumn{1}{c|}{} & \multicolumn{1}{c|}{\makecell[c]{MDD+\\
SCDA}}     & clp                  & inf                  & pnt                  & qdr                  & rel                  & skt                  & \multicolumn{1}{c|}{Avg.} \\ \hline
\multicolumn{1}{|c|}{clp}        & -                      &  19.3                         &  37.5                    &  11.1                      &  52.2                     &   41.0                   & \multicolumn{1}{c|}{32.2}     & \multicolumn{1}{c|}{} & \multicolumn{1}{c|}{clp}  &  -                  &  20.4                       & 36.6                       &  9.0                     &  50.7                     & 42.3                     & \multicolumn{1}{c|}{31.8}     & \multicolumn{1}{c|}{} & \multicolumn{1}{c|}{clp}  & -                      &  20.4                     & 43.3                   &  15.2                       & 59.3                      &  46.5                     & \multicolumn{1}{c|}{36.9}     \\
\multicolumn{1}{|c|}{inf}        & 30.2                     &   -                        & 31.2                      &  3.6                      & 44.0                       &  27.9                  & \multicolumn{1}{c|}{27.4}     & \multicolumn{1}{c|}{} & \multicolumn{1}{c|}{inf}  & 27.5                      & -                     &  25.7                      &  1.8                    &  34.7                      &  20.1                    & \multicolumn{1}{c|}{22.0}     & \multicolumn{1}{c|}{} & \multicolumn{1}{c|}{inf}  & 32.7                       & -                      & 34.5                     &   6.3                     & 47.6                     & 29.2                      & \multicolumn{1}{c|}{30.1}     \\
\multicolumn{1}{|c|}{pnt}        &   39.6                   &  18.7                          & -                     &  4.9                      & 54.5                      &   36.3                   & \multicolumn{1}{c|}{30.8}     & \multicolumn{1}{c|}{} & \multicolumn{1}{c|}{pnt}  &  42.6            &  20.0                     &  -                             &  2.5                    &  55.6                       & 38.5                    & \multicolumn{1}{c|}{31.8}     & \multicolumn{1}{c|}{} & \multicolumn{1}{c|}{pnt}  & 31.1                      & 6.6                      &  18.0                    & -                       & 28.8                      &  22.0                     & \multicolumn{1}{c|}{21.3}     \\
\multicolumn{1}{|c|}{qdr}        &  7.0                     &  0.9                        &  1.4                     &  -                      & 4.1                      &  8.3                    & \multicolumn{1}{c|}{4.3}     & \multicolumn{1}{c|}{} & \multicolumn{1}{c|}{qdr}  & 21.0                      &  4.5                   &  8.1                       &   -                    & 14.3                     & 15.7                      & \multicolumn{1}{c|}{12.7}     & \multicolumn{1}{c|}{} & \multicolumn{1}{c|}{qdr}  & 31.1                     &   6.6                     & 18.0                      &  -                     & 28.8                       & 22.0                      & \multicolumn{1}{c|}{21.3}     \\
\multicolumn{1}{|c|}{rel}        &  48.4                     &  22.2                         & 49.4                     &  6.4                      &  -                     &  38.8                    & \multicolumn{1}{c|}{33.0}     & \multicolumn{1}{c|}{} & \multicolumn{1}{c|}{rel}  &  51.9                 &  23.3                       & 50.4                        &  5.4                     & -                     & 41.4                      & \multicolumn{1}{c|}{34.5}     & \multicolumn{1}{c|}{} & \multicolumn{1}{c|}{rel}  & 55.5                      &  23.7                     & 52.9                      &  9.5                     &  -                     &  45.2                     & \multicolumn{1}{c|}{37.4}     \\
\multicolumn{1}{|c|}{skt}        &  46.9                   &  15.4                          &  37.0                      &  10.9                    &  47.0                     &  -                      & \multicolumn{1}{c|}{31.4}     & \multicolumn{1}{c|}{} & \multicolumn{1}{c|}{skt}  & 50.8                    & 20.3                        & 43.0                     & 2.9                      & 50.8                       & -                     & \multicolumn{1}{c|}{33.6}     & \multicolumn{1}{c|}{} & \multicolumn{1}{c|}{skt}  & 55.8                     &  20.1                      & 46.5                      &  15.0                     & 56.7                      &   -                    & \multicolumn{1}{c|}{38.8}     \\
\multicolumn{1}{|c|}{Avg.}       & 34.4                      &  15.3                        &  31.3                     &  7.4                      & 40.4                      &  30.5                    & \multicolumn{1}{c|}{26.6}     & \multicolumn{1}{c|}{} & \multicolumn{1}{c|}{Avg.} &  38.8                  & 17.7                        &  32.8                      &  4.3                     &  41.2                     &   31.6                   & \multicolumn{1}{c|}{27.7}     & \multicolumn{1}{c|}{} & \multicolumn{1}{c|}{Avg.} & 44.3                      &  18.1                     &  39.0                     & 10.8                      & 50.2                     &  37.2                      & \multicolumn{1}{c|}{33.3}     \\ \hline
\multicolumn{1}{l}{}             & \multicolumn{1}{l}{} & \multicolumn{1}{l}{}     & \multicolumn{1}{l}{} & \multicolumn{1}{l}{} & \multicolumn{1}{l}{} & \multicolumn{1}{l}{} & \multicolumn{1}{l}{}      & \multicolumn{1}{l}{}  & \multicolumn{1}{l}{}      & \multicolumn{1}{l}{} & \multicolumn{1}{l}{} & \multicolumn{1}{l}{} & \multicolumn{1}{l}{} & \multicolumn{1}{l}{} & \multicolumn{1}{l}{} & \multicolumn{1}{l}{}      & \multicolumn{1}{l}{}  & \multicolumn{1}{l}{}      & \multicolumn{1}{l}{} & \multicolumn{1}{l}{} & \multicolumn{1}{l}{} & \multicolumn{1}{l}{} & \multicolumn{1}{l}{} & \multicolumn{1}{l}{} & \multicolumn{1}{l}{}      \\ \hline
\multicolumn{1}{|c|}{ViT-B} & clp                  & \multicolumn{1}{c}{inf} & pnt                  & qdr                  & rel                  & skt                  & \multicolumn{1}{c|}{Avg.} & \multicolumn{1}{c|}{} & \multicolumn{1}{c|}{CD-Trans}     & clp                  & inf                  & pnt                  & qdr                  & rel                  & skt                  & \multicolumn{1}{c|}{Avg.} & \multicolumn{1}{c|}{} & \multicolumn{1}{c|}{\makecell[c]{Baseline\\-B}}     & clp                  & inf                  & pnt                  & qdr                  & rel                  & skt                  & \multicolumn{1}{c|}{Avg.} \\ \hline
\multicolumn{1}{|c|}{clp}        &  -                 &  27.2                            &   53.1                 &  13.2                         &  71.2                    &   53.3                   & \multicolumn{1}{c|}{43.6}     & \multicolumn{1}{c|}{} & \multicolumn{1}{c|}{clp}  &  -                    &  29.4                     & 57.2                     &  26.0                    & 72.6                    &  58.1                    & \multicolumn{1}{c|}{48.7}     & \multicolumn{1}{c|}{} & \multicolumn{1}{c|}{clp}  &     -                 &  30.9                    &  53.3                    &  16.3                    & 72.7                     & 55.4                     & \multicolumn{1}{c|}{45.7}     \\
\multicolumn{1}{|c|}{inf}        & 51.4                   & -                             & 49.3                     &  4.0                    & 66.3                        &   41.1                    & \multicolumn{1}{c|}{42.4}     & \multicolumn{1}{c|}{} & \multicolumn{1}{c|}{inf}  &  57.0                    & -                     & 54.4                     &  12.8                    & 69.5                     &  48.4                    & \multicolumn{1}{c|}{48.4}     & \multicolumn{1}{c|}{} & \multicolumn{1}{c|}{inf}  & 43.0                     &  -                    &   40.8                    &   7.8                    & 56.4                     & 35.9                     & \multicolumn{1}{c|}{36.8}     \\
\multicolumn{1}{|c|}{pnt}        & 53.1                  &  25.6                            & -                &  4.8                            &  70.0                    &  41.8                     & \multicolumn{1}{c|}{39.1}     & \multicolumn{1}{c|}{} & \multicolumn{1}{c|}{pnt}  & 62.9                     & 27.4                     & -                     &  15.8                    & 72.1                     &  53.9                    & \multicolumn{1}{c|}{46.4}     & \multicolumn{1}{c|}{} & \multicolumn{1}{c|}{pnt}  &  55.7                    & 28.6                     & -                     & 7.4                     & 70.5                     & 48.3                     & \multicolumn{1}{c|}{42.1}     \\
\multicolumn{1}{|c|}{qdr}        &  30.5                  &   4.5                         &  16.0               &   -                       &  27.0                         &   19.3                   & \multicolumn{1}{c|}{19.5}     & \multicolumn{1}{c|}{} & \multicolumn{1}{c|}{qdr}  &  44.6                    &  8.9                    &  29.0                    & -                     &  42.6                     & 28.5                      & \multicolumn{1}{c|}{30.7}     & \multicolumn{1}{c|}{} & \multicolumn{1}{c|}{qdr}  &  25.5                    & 5.2                     &  9.7                     &  -                    &  15.5                     &  17.1                    & \multicolumn{1}{c|}{14.6}     \\
\multicolumn{1}{|c|}{rel}        & 58.4                &  29.0                             & 60.0                 &  6.0                           &   -                    &    45.8                    & \multicolumn{1}{c|}{39.9}     & \multicolumn{1}{c|}{} & \multicolumn{1}{c|}{rel}  & 66.2                    &  31.0                       &  61.5                   &   16.2                      &  -                     &   52.9                    & \multicolumn{1}{c|}{45.6}     & \multicolumn{1}{c|}{} & \multicolumn{1}{c|}{rel}  &  62.3                    &  32.5                    & 62.5                     &  8.2                    & -                    &  50.7                    & \multicolumn{1}{c|}{43.2}     \\
\multicolumn{1}{|c|}{skt}        &  63.9           & 23.8                          &  52.3                              &   14.4                    &  67.4                      & -                      & \multicolumn{1}{c|}{44.4}     & \multicolumn{1}{c|}{} & \multicolumn{1}{c|}{skt}  & 69.0             &  29.6                            &    59.0                 &  27.2                        &    72.5                    &   -                   & \multicolumn{1}{c|}{51.5}     & \multicolumn{1}{c|}{} & \multicolumn{1}{c|}{skt}  & 66.4                     & 30.6                     & 58.0                     &  18.1                    & 70.1                     & -                     & \multicolumn{1}{c|}{48.6}     \\
\multicolumn{1}{|c|}{Avg.}       & 51.5                 &   22.0                       &  46.1                           &  8.5                     &  60.4                    & 40.3                     & \multicolumn{1}{c|}{38.1}     & \multicolumn{1}{c|}{} & \multicolumn{1}{c|}{Avg.} &  59.9                   &  25.3                      & 52.2                      &  19.6                      &  65.9                     &  48.4                     & \multicolumn{1}{c|}{45.2}     & \multicolumn{1}{c|}{} & \multicolumn{1}{c|}{Avg.} &  50.6                    &  25.6                    & 44.9                     & 11.6                     & 57.0                      & 41.5                     & \multicolumn{1}{c|}{38.5}     \\ \hline
\multicolumn{1}{l}{}             & \multicolumn{1}{l}{} & \multicolumn{1}{l}{}     & \multicolumn{1}{l}{} & \multicolumn{1}{l}{} & \multicolumn{1}{l}{} & \multicolumn{1}{l}{} & \multicolumn{1}{l}{}      & \multicolumn{1}{l}{}  & \multicolumn{1}{l}{}      & \multicolumn{1}{l}{} & \multicolumn{1}{l}{} & \multicolumn{1}{l}{} & \multicolumn{1}{l}{} & \multicolumn{1}{l}{} & \multicolumn{1}{l}{} & \multicolumn{1}{l}{}      & \multicolumn{1}{l}{}  & \multicolumn{1}{l}{}      & \multicolumn{1}{l}{} & \multicolumn{1}{l}{} & \multicolumn{1}{l}{} & \multicolumn{1}{l}{} & \multicolumn{1}{l}{} & \multicolumn{1}{l}{} & \multicolumn{1}{l}{}      \\ \hline
\multicolumn{1}{|c|}{\makecell[c]{Baseline\\-B+Ours}} & clp                  & inf                      & pnt                  & qdr                  & rel                  & skt                  & \multicolumn{1}{c|}{Avg.} & \multicolumn{1}{c|}{} & \multicolumn{1}{c|}{SSRT-B}     & clp                  & inf                  & pnt                  & qdr                  & rel                  & skt                  & \multicolumn{1}{c|}{Avg.} & \multicolumn{1}{c|}{} & \multicolumn{1}{c|}{\makecell[c]{SSRT-B\\+Ours}}     & clp                  & inf                  & pnt                  & qdr                  & rel                  & skt                  & \multicolumn{1}{c|}{Avg.} \\ \hline
\multicolumn{1}{|c|}{clp}        &  -                    & 31.2                        &  60.5                    &  29.6                   &  77.8                    &  61.8                    & \multicolumn{1}{c|}{\cellcolor[gray]{0.9}52.2}     & \multicolumn{1}{c|}{} & \multicolumn{1}{c|}{clp}  &  -                    & 33.8                    & 60.2                     & 19.4                     & 75.8                      &  59.8                    & \multicolumn{1}{c|}{49.8}     & \multicolumn{1}{c|}{} & \multicolumn{1}{c|}{clp}  &   -                   &    33.5                 & 63.1                      &   28.5                   &  78.5                    &  63.4                    & \multicolumn{1}{c|}{\textcolor{red}{\textbf{\cellcolor[gray]{0.9}53.4}}}     \\
\multicolumn{1}{|c|}{inf}        &  61.5                    &   -                       &  59.7                    &  16.8                   & 76.2                     & 54.2                     & \multicolumn{1}{c|}{\cellcolor[gray]{0.9}53.7}     & \multicolumn{1}{c|}{} & \multicolumn{1}{c|}{inf}  & 55.5                      &   -                     & 54.0                     &  9.0                    & 68.2                     & 44.7                     & \multicolumn{1}{c|}{46.3}     & \multicolumn{1}{c|}{} & \multicolumn{1}{c|}{inf}  & 65.3                    &  -                  &  61.2                    &      15.0                 &  77.1                     &    56.0                  & \multicolumn{1}{c|}{\textcolor{red}{\textbf{\cellcolor[gray]{0.9}54.9}}}     \\
\multicolumn{1}{|c|}{pnt}        &  63.8                    &  29.9                       &  -                  &  16.8                    &  77.1                    &  56.8                    & \multicolumn{1}{c|}{\cellcolor[gray]{0.9}48.9}     & \multicolumn{1}{c|}{} & \multicolumn{1}{c|}{pnt}  &  61.7                    &  28.5                    &   -                   &  8.4                     & 71.4                     &  55.2                    & \multicolumn{1}{c|}{45.0}     & \multicolumn{1}{c|}{} & \multicolumn{1}{c|}{pnt}  & 66.3                     & 31.1                     &  -                    &    15.5                & 76.7                     &  59.3                    & \multicolumn{1}{c|}{\textcolor{red}{\textbf{\cellcolor[gray]{0.9}49.8}}}     \\
\multicolumn{1}{|c|}{qdr}        &  36.0                   &  6.7                        &   14.9                   &  -                    & 30.6                     &  27.0                    & \multicolumn{1}{c|}{\cellcolor[gray]{0.9}23.0}     & \multicolumn{1}{c|}{} & \multicolumn{1}{c|}{qdr}  & 42.5                     &  8.8                    & 24.2                     & -                     &  37.6                    & 33.6                     & \multicolumn{1}{c|}{29.3}     & \multicolumn{1}{c|}{} & \multicolumn{1}{c|}{qdr}  & 53.3                     &  8.5                   &  28.2                    & -                     & 38.2                     &  39.5                    & \multicolumn{1}{c|}{\textcolor{red}{\textbf{\cellcolor[gray]{0.9}33.5}}}     \\
\multicolumn{1}{|c|}{rel}        & 68.0                    & 32.5                         & 64.7                     &  19.3                   & -                     & 58.4                     & \multicolumn{1}{c|}{\cellcolor[gray]{0.9}48.6}     & \multicolumn{1}{c|}{} & \multicolumn{1}{c|}{rel}  & 69.9                     & 37.1                     & 66.0                     & 10.1                   &  -                    &   58.9                    & \multicolumn{1}{c|}{48.4}     & \multicolumn{1}{c|}{} & \multicolumn{1}{c|}{rel}  &  72.3                    & 36.6                     & 66.5                     &  20.0                    &  -                    & 61.9                     & \multicolumn{1}{c|}{\textcolor{red}{\textbf{\cellcolor[gray]{0.9}51.5}}}     \\
\multicolumn{1}{|c|}{skt}        & 71.1                     & 31.9                         &  64.0                    & 30.1                     & 77.5                    & -                     & \multicolumn{1}{c|}{\cellcolor[gray]{0.9}54.9}     & \multicolumn{1}{c|}{} & \multicolumn{1}{c|}{skt}  & 70.6                      & 32.8                     & 62.2                       & 21.7                      &    73.2                  &  -                     & \multicolumn{1}{c|}{52.1}     & \multicolumn{1}{c|}{} & \multicolumn{1}{c|}{skt}  &      73.5                &  32.9                    & 65.7                     &  28.1                    &  78.1                    &  -                    & \multicolumn{1}{c|}{\textcolor{red}{\textbf{\cellcolor[gray]{0.9}55.7}}}     \\
 \multicolumn{1}{|c|}{\cellcolor[gray]{0.9}Avg.}       &  \cellcolor[gray]{0.9} 60.1                    &\cellcolor[gray]{0.9} 26.4                         &      \cellcolor[gray]{0.9}  52.8              &  \cellcolor[gray]{0.9} 22.5                   & \cellcolor[gray]{0.9} 67.8                    &  \cellcolor[gray]{0.9}51.6                    & \multicolumn{1}{c|}{\cellcolor[gray]{0.9}46.9}     & \multicolumn{1}{c|}{} & \multicolumn{1}{c|}{Avg.} &  60.0                    &  28.2                      &  53.3                     & 13.7                     & 65.3                      &  50.4                      & \multicolumn{1}{c|}{45.2}     & \multicolumn{1}{c|}{} & \multicolumn{1}{c|}{\cellcolor[gray]{0.9}Avg.} &  \cellcolor[gray]{0.9}\textcolor{red}{\textbf{66.1}}                    &\cellcolor[gray]{0.9} \textcolor{red}{\textbf{28.5}}                     &\cellcolor[gray]{0.9} \textcolor{red}{\textbf{56.9}}                     & \cellcolor[gray]{0.9}\textcolor{red}{\textbf{21.4}}                     &\cellcolor[gray]{0.9} \textcolor{red}{\textbf{69.7}}                     & \cellcolor[gray]{0.9}\textcolor{red}{\textbf{56.0}}                     & \multicolumn{1}{c|}{\textcolor{red}{\textbf{\cellcolor[gray]{0.9}49.8}}}     \\ \hline
\end{tabular}}
\label{tab:domainnet}
\end{table*}

\subsubsection{Implementation}
We set two baselines for our validation, \ieno, Baseline-B and Baseline-S. Here, Baseline-B/Baseline-S are the ViT-B/ViT-S~\cite{dosovitskiy2020imageViT} backbones with adversarial alignment of DANN~\cite{ganin2016domain_dann}, respectively.   
To demonstrate the applicability of our method, we further incorporate our method into the state-of-the-art scheme SSRT~\cite{sun2022safeSSRT}. 
Concretely, SSRT~\cite{sun2022safeSSRT} proposes the safe-training and multi-layer target perturbation for Transformer-based UDA, which achieves optimal performance in OfficeHome~\cite{venkateswara2017deepOffice-Home}, VisDA-2017~\cite{peng2017visdaViSDA2017} and DomainNet~\cite{peng2019momentDomainNet}.  We adopt the  Stochastic Gradient Descent algorithm with the momentum of 0.9 and weight decay 1e-4 as our optimizer. The initial learning rate is 1e-3 for DomainNet and OfficeHome, 5e-5 for VisDA-2017. The batchsize is 32 for each dataset. 


\subsection{Comparisons with the State-of-the-arts}
\label{sec:sota}
We experimentally validate our method on three commonly-used benchmarks for UDA, \ieno, DomainNet~\cite{peng2019momentDomainNet}, Office-Home~\cite{venkateswara2017deepOffice-Home}, VisDA-2017~\cite{peng2017visdaViSDA2017}.  To validate the effectiveness of our proposed SAMB, we select CNN-based and Transformer-based methods for comparison. For CNN-based methods, we compare ours with ResNet-101~\cite{he2016deepResNet}, ResNet-50~\cite{he2016deepResNet}, DANN~\cite{ganin2016domain_dann}, CDAN~\cite{long2018conditionalCDAN}, SAFN~\cite{xu2019largerSAFN}, SWD~\cite{lee2019slicedSWD}, and SHOT~\cite{liang2020weSHOT}. We also compare our SAMB with the state-of-the-art Transformer-based UDA works, including CDTrans~\cite{xu2021cdtransCDTrans}, BCAT-DTF~\cite{wang2022domainBCAT}, WinTR~\cite{ma2021exploitingWinTR}, and TVT~\cite{yang2021tvtTVT}. Among them, \textbf{CDTrans}~\cite{xu2021cdtransCDTrans} exploits the cross-attention of different domains to learn the domain invariant knowledge for alignment. \textbf{BCAT}~\cite{wang2022domainBCAT} improves the CDTrans with a quadruple-branch transformer by adding the dual cross-attention between the source and target domain. \textbf{WinTR}~\cite{ma2021exploitingWinTR} utilizes the task-specific classifier and masked self-attention to preserve the domain-specific and domain-invariant knowledge, thereby achieving the win-win Transformer-based UDA. \textbf{TVT}~\cite{yang2021tvtTVT} introduces the transferability adaption module to implement the fine-grained feature alignment for UDA. 

\begin{table*}[htp]
\centering
\caption{Ablation studies on different message passing.}
\setlength{\tabcolsep}{1.7mm}{\begin{tabular}{cccccccccccccc}
\hline
Method                              & Ar$\xrightarrow{}$Cl                & Ar$\xrightarrow[]{}$Pr                                 & Ar$\xrightarrow[]{}$Rw   & Cl$\xrightarrow[]{}$Ar                                  & Cl$\xrightarrow[]{}$Pr   & Cl$\xrightarrow[]{}$Rw    & Pr$\xrightarrow[]{}$Ar  & Pr$\xrightarrow[]{}$Cl    &  Pr$\xrightarrow[]{}$Rw     & Rw$\xrightarrow[]{}$Ar  & Rw$\xrightarrow[]{}$Cl  & Rw$\xrightarrow[]{}$Pr  & Avg.                                  \\ \hline
Baseline & 62.88 & 79.52 & 86.09 & 77.62 & 81.91 & 82.96 & 75.40 & 63.13 & 87.53 & 80.63 & 65.22 & 87.11 & 77.50 \\
SAMG & 61.48 & 79.59 & 85.31 & 77.42 & 81.03 & 83.11 & 74.17  & 60.94 & 86.96 & 80.14 & 63.94 & 86.75 & 76.74 \\ 
G-G & 62.50 & 80.74 & 85.43 & 78.20 & 81.66 & 83.31 & 75.86 & 61.67 & 87.08 & 80.14 & 64.58 & 86.71 & 77.32 \\ 
G-L & 62.66 & 80.85 & 85.61 & 77.92 & 81.50 & 83.54 & 76.18 & 61.79 & 87.38 & 80.68 & 65.06 & 87.05 & 77.52  \\ 
SAMB & 63.02 & 80.85 & \tcrt{86.21} & \tcrt{79.03} & 82.06 & \tcrt{84.00} & 76.01  & 62.72 & 87.74 & 81.29 & 64.92 & 87.41 & 77.94 \\ 
SAMG-D & 62.58 & 81.23 & 85.75 & 77.92 & 81.93 & 83.93 & 76.89  & 61.56 & 87.56 & 80.02 & 63.78 & 87.45 & 77.55\\ 
G-L-D & 63.05 & 80.33 &  85.06 & 77.79 &  81.35 & 83.34 & 75.48 & 61.88 & 87.15 & 79.73 & 63.94 & 87.00 & 77.18 \\
\rowcolor[gray]{0.9} SAMB-D & \tcrt{63.85} & \tcrt{82.05} & 86.14 & 77.87 & \tcrt{82.52} & 83.41 & \tcrt{76.35} & \tcrt{63.62} & \tcrt{87.81} & \tcrt{81.62} & \tcrt{65.43} & \tcrt{87.81} & \tcrt{78.21} \\ \hline
\end{tabular}}
\label{tab:different_message}
\end{table*}
In particular, we incorporate our SAMB into three baselines, \ieno, Baseline-B, and Baseline-S, and a state-of-the-art Transformer-UDA work SSRT~\cite{sun2022safeSSRT}. Here, Baseline-S and Baseline-B refer to the backbones ViT-S/ViT-B with the adversarial-based alignment of DANN~\cite{ganin2016domain_dann}. SSRT~\cite{sun2022safeSSRT} introduces the self-refinement loss and feature perturbation for better domain alignment. As shown in Table.~\ref{tab:visda-2017}, on VisDA-2017~\cite{peng2017visdaViSDA2017}, ours improves the Baseline-S, Baseline-B, and SSRT-B by a large margin of 3.93\%, 2.10\%, and 1.65\%, respectively. It is noteworthy that the SSRT-B combined with ours can achieve 90.41\% for the target domain, which outperforms all recent transformer-based methods, \egno, BCAT-DTF~\cite{wang2022domainBCAT}, WinTR-B~\cite{ma2021exploitingWinTR}, TVT~\cite{yang2021tvtTVT}, and SSRT-B~\cite{sun2022safeSSRT}. For Office-Home~\cite{venkateswara2017deepOffice-Home} in Table~\ref{tab:office-home}, ours can improve the Baseline-S, Baseline-B, SSRT-S, SSRT-B by  2.31\%, 0.93\%, 1.45\%, and 0.77\%. Moreover, with SSRT-B as the backbone, our SAMB can exceed the TVT by 2.64\%.  
Despite the most challenging dataset DomainNet~\cite{peng2019momentDomainNet}, containing a large domain gap between different domains, ours shows great superiority compared with other methods. From the table~\ref{tab:domainnet}, by incorporating our SAMB, Baseline-B can improve itself 8.4\% on average, of which the total accuracy 46.9\% outperforms the SOTA method SSRT-B~\cite{sun2022safeSSRT} by a large margin of 1.7\%. We also incorporate our method into SSRT-B, which exceeds the SSRT-B by a gain of 4.6\%, and achieves the optimal average accuracy on the target domain with 49.8\%, despite SSRT being already strong for UDA. 

\begin{table*}[htp]
\centering
\caption{Ablation studies on the number of group tokens with the Dynamic Semantic-aware Message Broadcasting.}
\setlength{\tabcolsep}{1.7mm}{\begin{tabular}{cccccccccccccc}
\hline
Numbers                              & Ar$\xrightarrow{}$Cl                & Ar$\xrightarrow[]{}$Pr                                 & Ar$\xrightarrow[]{}$Rw   & Cl$\xrightarrow[]{}$Ar                                  & Cl$\xrightarrow[]{}$Pr   & Cl$\xrightarrow[]{}$Rw    & Pr$\xrightarrow[]{}$Ar  & Pr$\xrightarrow[]{}$Cl    &  Pr$\xrightarrow[]{}$Rw     & Rw$\xrightarrow[]{}$Ar  & Rw$\xrightarrow[]{}$Cl  & Rw$\xrightarrow[]{}$Pr  & Avg.                                  \\ \hline
2 & 62.45 & 81.46 & 85.84 & \tcrt{78.86} & 81.82 & \tcrt{84.07} & 76.72 & 62.27 & 87.63 & 81.25 & 64.95 & 87.61 & 77.91 \\
\rowcolor[gray]{0.9} 4 & 63.48 & \tcrt{82.05} & \tcrt{86.14} & 77.87 & 82.52 & 83.41 & 76.35  & \tcrt{63.62} & 87.81 & \tcrt{81.62} & 65.43 & \tcrt{87.81} & \tcrt{78.21} \\ 
8 & 63.02 & 81.62 & 85.82 & 78.25 & \tcrt{82.92} & 83.77 & 76.68 &  62.68 & 87.74 & 80.92 & 65.16 & 87.75 &  78.03  \\ 
16 & \tcrt{63.83} & 81.89 & 86.00 & 78.16 & 82.45 & 83.68 & \tcrt{77.50} & 62.93 & 87.77 & 81.25 & 64.88 & 87.45 & 78.15  \\ 
32 & 63.18 & 81.42 & 86.02 & 78.37 & 82.54 & 83.29 & 76.97  & 63.39 & \tcrt{87.95} & 80.92 & \tcrt{65.89} & 87.75 & 78.14 \\ \hline
\end{tabular}}
\label{tab:number}
\end{table*}

\begin{figure}[htp]
    \centering
    \includegraphics[width=0.8\linewidth]{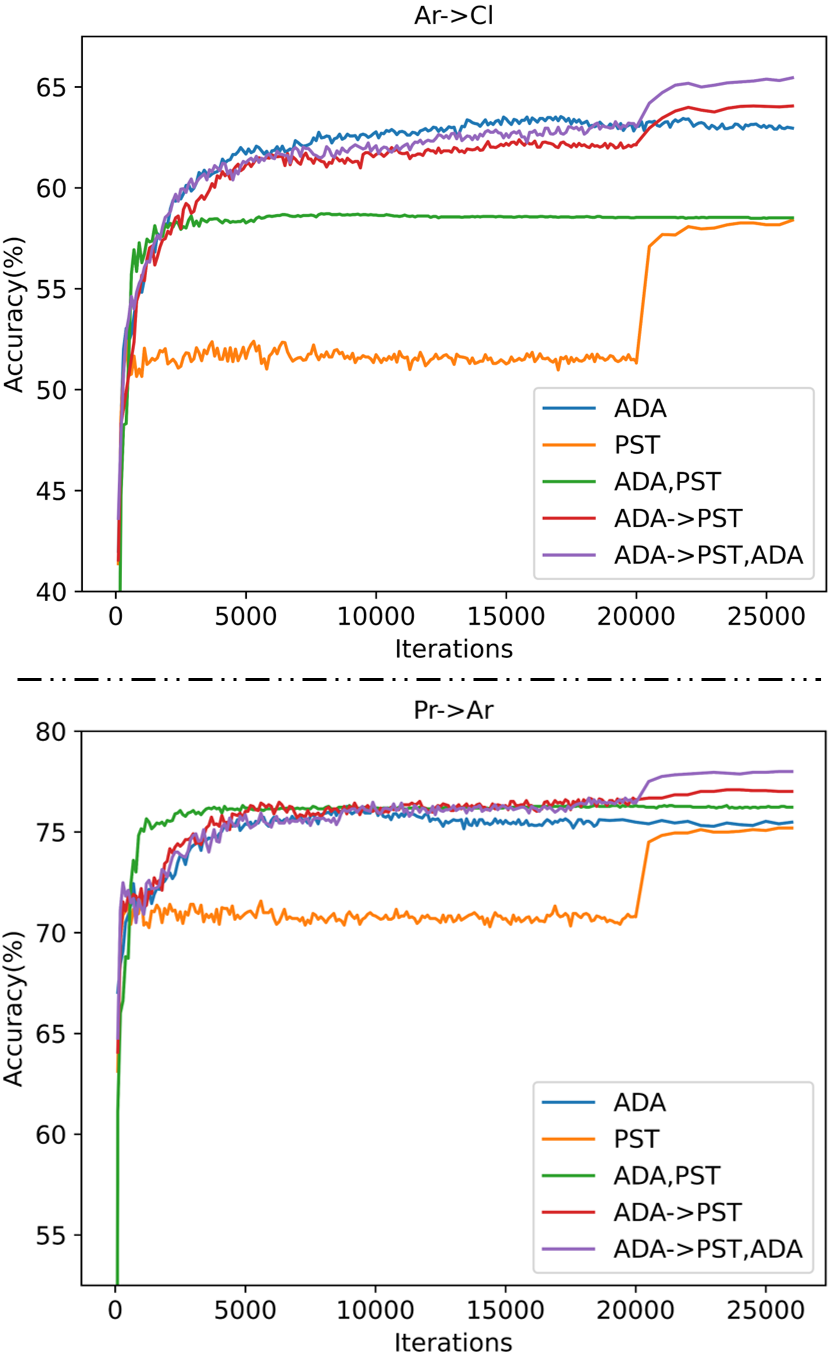}
    \caption{A comparison for training process between different training mechanisms. ADA$\xrightarrow[]{}$PST denotes that training the network with adversarial-based feature alignment and then with pseudo-label based self-training. }
    \label{fig:curve_training_mechanism}
\end{figure}
\subsection{Ablation Study}
\label{sec:ab}
\subsubsection{Effects of dynamic semantic-aware message broadcasting}
\label{sec:ab_ms}
We conduct extensive ablation studies to compare the different message-passing strategies, which can validate the effectiveness of our proposed SAMB. Here, we adopt the ViT-small~\cite{dosovitskiy2020imageViT} with DANN~\cite{ganin2016domain_dann} as the baseline. As shown in Table~\ref{tab:different_message}, we
delicately design the  message-passing strategies for comparison, \ieno, SAMA, SAMA-D, SAMB, SAMB-D, G-G, G-L, G-L-D. 1) ``SAMB-D" denotes our dynamic semantic-aware message broadcasting used in our paper, where each group token aggregates global information from all image tokens but distributes to different image tokens dynamically. 2) ``SAMB" refers to ``SAMB-D" without our dynamic assignment. 3) ``SAMG", as an inverse strategy of ``SAMB", is semantic-aware message aggregation, where each region token aggregate messages from different image tokens but distribute them to all image token. 4) ``SAMG-D" denotes the ``SAMG" with our dynamic assignment. 
5) In the ``G-G" setting, each group token interacts with messages with all image tokens, we can regard them as multiple class tokens. 6) In the ``G-L" setting, each group token is responsible for the local-to-local message passing. We retain the class token for the global-to-global message passing in this setting. 7) For ``G-L-D", we replace the local-to-local message passing in ``G-L" with dynamic semantic-aware message aggregation and dynamic semantic-aware message broadcasting. 
We can observe that the ``SAMG" and  ``G-G" do not work for Transformer UDA, which causes a noticeable performance drop, since they change the message scale a lot. The ``G-L" strategy changes the message scale slightly and introduces the semantic-aware message passing to the different local regions, which only brings a gain of 0.02\%. ``G-L-D" further destroys the stability of the message scale for each image token in the message aggregation process, which causes the performance drop. Compared with the above message-passing methods, our SAMB and SAMB-D achieve the best performance by a gain of 0.44\% and 0.71\%, which reveals the superiority of our scheme. The reasons are as follows: 1) Our proposed semantic-aware message broadcasting does not change the message scale in the message-passing process. 2) Our SAMB-D enables more informative and diverse features for domain alignment by semantic-aware message broadcasting.  
We also study the optimal number of region tokens. As shown in Table~\ref{tab:number}, we set the number of group tokens as 2, 4, 8, 16, and 32, respectively. Experiments reveal that the number $4$ is enough for transformer UDA in classification, which is used in our paper. 

\begin{table*}[htp]
\centering
\caption{Ablation studies on different training mechanisms.}
\setlength{\tabcolsep}{1.7mm}{\begin{tabular}{cccccccccccccc}
\hline
Method                              & Ar$\xrightarrow{}$Cl                & Ar$\xrightarrow[]{}$Pr                                 & Ar$\xrightarrow[]{}$Rw   & Cl$\xrightarrow[]{}$Ar                                  & Cl$\xrightarrow[]{}$Pr   & Cl$\xrightarrow[]{}$Rw    & Pr$\xrightarrow[]{}$Ar  & Pr$\xrightarrow[]{}$Cl    &  Pr$\xrightarrow[]{}$Rw     & Rw$\xrightarrow[]{}$Ar  & Rw$\xrightarrow[]{}$Cl  & Rw$\xrightarrow[]{}$Pr  & Avg.                                  \\ \hline
ADA & 62.88 & 79.52 & 86.09 & 77.62 & 81.91 & 82.96 & 75.4 & 63.13 & 87.53 & 80.63 & 65.22 & 87.11 & 77.50 \\
PST & 57.91 & \tcrt{85.19} & \tcrt{87.65} & 79.47 & \tcrt{86.12} & \tcrt{86.79} & 75.26 & 54.56 & 87.25 & 78.93 & 54.15 & 87.67 & 76.75 \\ 
ADA,PST & 58.39 & 83.62 & 87.64 & 79.62 & 84.94 & 86.75 & 76.27 & 57.43 & 88.51 & 79.62 & 58.95 & 87.25 & 77.44 \\ 
ADA$\xrightarrow{}$PST & 64.72 & 84.46 & 87.12 & \tcrt{80.30} & 85.11 & 85.63 & 77.22 & 63.46 & 88.59 & \tcrt{82.16} & 67.03 & 87.99 & 79.48 \\ 
\rowcolor[gray]{0.9} ADA$\xrightarrow[]{}$PST,ADA & \tcrt{64.65} & 84.07 & 86.99 & 80.76 & 84.68 & 85.72 & \tcrt{78.99} & \tcrt{63.39} & \tcrt{88.82} & 81.21 & \tcrt{67.01} & \tcrt{88.06} & \tcrt{79.53} \\
   \hline
\end{tabular}}
\label{tab:trainingmechanism}
\end{table*}

\begin{figure}[htp]
    \centering
    \includegraphics[width=1.0\linewidth]{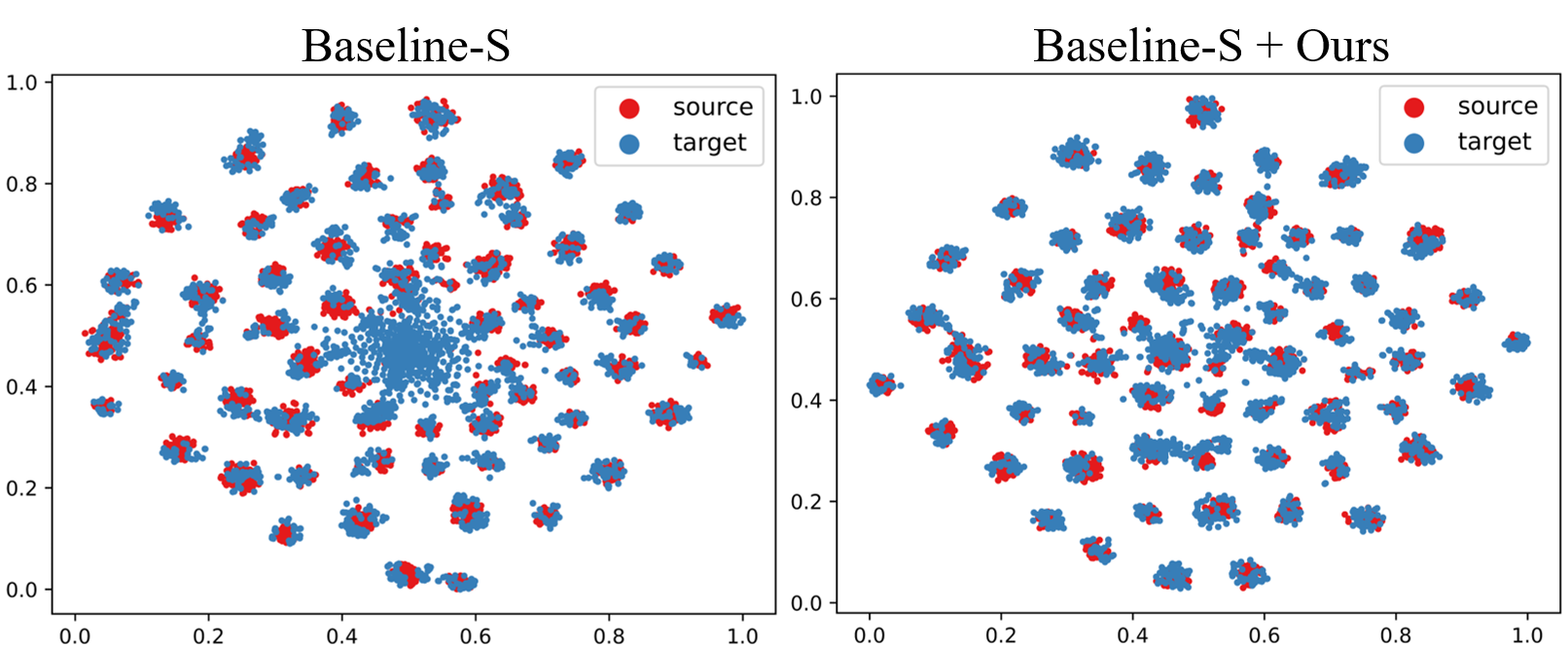}
    \caption{t-SNE visualization of features learned by Baseline-S (left) and Baseline-S+Ours (right). }
    \label{fig:tsne-vis}
    \vspace{-3mm}
\end{figure}
\subsubsection{Effects of different training mechanisms}
It is noteworthy that recent works on UDA are mostly  based on two training strategies, \ieno, adversarial-based feature alignment (ADA), and pseudo-label based self-training (PST). However, few works explore how to combine them in the training process. In this paper, we systematically investigate different training mechanisms for Transformer UDA, including adversarial-based feature alignment (\ieno, ADA), pseudo-label based self-training (\ieno, PST), and different combination 
strategies of the above two training mechanisms. For the combination of ADA and PST, a simple scheme is to optimize the network with ADA and PST together, which is represented as ``ADA, PST". We also attempt to optimize the network with ADA and PST in a sequential manner (\ieno, ``ADA$\xrightarrow[]{}$PST"). It means we first conduct domain alignment for the vision transformer with domain adversarial loss in Eq.~\ref{eq:ada_uda} and then optimize the network with PST. Apart from that, we also attempt to substitute the PST in the second stage of ``ADA$\xrightarrow[]{}$PST" with the combination of ADA and PST (\ieno, ``ADA,PST"), which is denoted as ``ADA$\xrightarrow[]{}$PST,ADA".

Based on the experimental results, we can have the following findings and conclusions:
1) As shown in Table~\ref{tab:trainingmechanism}, ADA achieves a more promising performance than PST by a gain of 0.75\% for the vision transformer. 2) The simple combination of ADA and PST (\ieno,  ``ADA,PST") does not bring a performance improvement compared with ADA. To analyze the reason for the above experimental results, we also visualize the training process in Fig.~\ref{fig:curve_training_mechanism}. From the figure, we can observe that the integration of PST to ADA causes a more fast convergence compared with ADA, but brings a sub-optimal solution for the vision transformer. 3) Notably, ``ADA$\xrightarrow[]{}$PST,ADA" achieves a gain of 1.98\% compared with only ADA, which reveals that self-training is complementary to adversarial-based alignment. From the Fig.~\ref{fig:curve_training_mechanism}, we can find that adversarial-based alignment provides a better initialization for PST, which brings better performance. 4) Lastly, the best training mechanism is the ``ADA$\xrightarrow[]{}$PST,ADA", which outperforms the ADA by 2.03\% and PST by 2.78\%. This reveals that adversarial-based alignment can further bring performance improvement in the second stage combined with PST.   




\subsection{Visualization}
\label{sec:vis}
To learn about whether our SAMB can achieve semantic-aware message broadcasting, we visualize the assignment of each group token in Fig.~\ref{fig:region_assignment}. From the table, we can observe that each group token is responsible for the message broadcasting of different semantics. For example, in the first line, the mountains, birds, and background are assigned to different group tokens, which means our SAMB learns the semantic-adaptive message broadcasting. This enables the features learned by different group tokens to be more informative and diverse.  Domain alignment with such features is more flexible and effective. To investigate the characteristics of the alignment feature, we compare the t-SNE visualizations of the Baseline-S and  our proposed SAMB on Pr$\xrightarrow[]{}$Cl of OfficeHome~\cite{venkateswara2017deepOffice-Home}  n Fig.~\ref{fig:tsne-vis}. From the figure, we can find that our method let the features in different clusters be more diverse while letting the alignment of the source and target domains be more compact, which also proves our motivation and conclusion.  

\subsection{Complexity Analysis}
\label{sec:complexity}
In this section, we discuss the complexity introduced by the group tokens. As shown in Table~\ref{tab:complexity}, our proposed SAMB only increases the parameters by approximately 0.4\% for Baseline-S, and 0.6\% for Baseline-B, which is very tiny for the vision transformer. For computational complexity (\ieno, Flops), ours only cause a slight increase of 1.28\% for Baseline-S and 1.32\%. This shows that our SAMB is efficient but effective for Transformer-based unsupervised domain adaptation.


\begin{table}[]
\caption{Complexity Analysis for our proposed SAMB.}
\label{tab:complexity}
\setlength{\tabcolsep}{0.75mm}{\begin{tabular}{c|c|c|c|c}
\hline
Methods        & Baseline-S & Baseline-S+Ours & Baseline-B & Baseline-B+Ours \\ \hline
Parameters (M) & 31.37      & 31.52           & 96.68      & 97.27           \\ \hline
Flops (G)      & 4.613      & 4.676           & 17.588     & 17.821          \\ \hline
\end{tabular}}
\end{table}

\section{Conclusion}
\label{sec:conlusion}
 In this paper, we propose the novel Semantic-aware Message Broadcasting (SAMB) for the Transformer-based UDA, which enables more informative and flexible feature alignment and improves domain adaptability. Particularly, we introduce a group of learned tokens and encourage them to learn more diverse and rich information based on the differences in regional semantics through SAMB. For the optimization, we revisit two popular training mechanisms, \ieno, Adversarial-based feature alignment (ADA), and Pseudo-label based self-training (PST). Based on the abundant experiments, we propose a simple but effective training paradigm ``ADA$\xrightarrow[]{}$PST, ADA", which achieves more promising domain adaptability. Extensive experiments on multiple benchmarks on UDA have demonstrated the effectiveness of our proposed method.

\bibliographystyle{IEEEtran}
\bibliography{references}

\begin{thebibliography}{10}
\providecommand{\url}[1]{#1}
\csname url@samestyle\endcsname
\providecommand{\newblock}{\relax}
\providecommand{\bibinfo}[2]{#2}
\providecommand{\BIBentrySTDinterwordspacing}{\spaceskip=0pt\relax}
\providecommand{\BIBentryALTinterwordstretchfactor}{4}
\providecommand{\BIBentryALTinterwordspacing}{\spaceskip=\fontdimen2\font plus
\BIBentryALTinterwordstretchfactor\fontdimen3\font minus
  \fontdimen4\font\relax}
\providecommand{\BIBforeignlanguage}[2]{{%
\expandafter\ifx\csname l@#1\endcsname\relax
\typeout{** WARNING: IEEEtran.bst: No hyphenation pattern has been}%
\typeout{** loaded for the language `#1'. Using the pattern for}%
\typeout{** the default language instead.}%
\else
\language=\csname l@#1\endcsname
\fi
#2}}
\providecommand{\BIBdecl}{\relax}
\BIBdecl

\bibitem{dosovitskiy2020imageViT}
A.~Dosovitskiy, L.~Beyer, A.~Kolesnikov, D.~Weissenborn, X.~Zhai,
  T.~Unterthiner, M.~Dehghani, M.~Minderer, G.~Heigold, S.~Gelly \emph{et~al.},
  ``An image is worth 16x16 words: Transformers for image recognition at
  scale,'' \emph{arXiv preprint arXiv:2010.11929}, 2020.

\bibitem{carion2020endDETR}
N.~Carion, F.~Massa, G.~Synnaeve, N.~Usunier, A.~Kirillov, and S.~Zagoruyko,
  ``End-to-end object detection with transformers,'' in \emph{European
  conference on computer vision}.\hskip 1em plus 0.5em minus 0.4em\relax
  Springer, 2020, pp. 213--229.

\bibitem{he2017maskMASKRCNN}
K.~He, G.~Gkioxari, P.~Doll{\'a}r, and R.~Girshick, ``Mask r-cnn,'' in
  \emph{Proceedings of the IEEE international conference on computer vision},
  2017, pp. 2961--2969.

\bibitem{redmon2016youYOLO}
J.~Redmon, S.~Divvala, R.~Girshick, and A.~Farhadi, ``You only look once:
  Unified, real-time object detection,'' in \emph{Proceedings of the IEEE
  conference on computer vision and pattern recognition}, 2016, pp. 779--788.

\bibitem{cheng2022masked}
B.~Cheng, I.~Misra, A.~G. Schwing, A.~Kirillov, and R.~Girdhar,
  ``Masked-attention mask transformer for universal image segmentation,'' in
  \emph{Proceedings of the IEEE/CVF Conference on Computer Vision and Pattern
  Recognition}, 2022, pp. 1290--1299.

\bibitem{li2022exploring}
Y.~Li, H.~Mao, R.~Girshick, and K.~He, ``Exploring plain vision transformer
  backbones for object detection,'' \emph{arXiv preprint arXiv:2203.16527},
  2022.

\bibitem{chu2021twinsTwins}
X.~Chu, Z.~Tian, Y.~Wang, B.~Zhang, H.~Ren, X.~Wei, H.~Xia, and C.~Shen,
  ``Twins: Revisiting the design of spatial attention in vision transformers,''
  \emph{Advances in Neural Information Processing Systems}, vol.~34, pp.
  9355--9366, 2021.

\bibitem{zhou2022domainSurveyKaiyang}
K.~Zhou, Z.~Liu, Y.~Qiao, T.~Xiang, and C.~C. Loy, ``Domain generalization: A
  survey,'' \emph{IEEE Transactions on Pattern Analysis and Machine
  Intelligence}, 2022.

\bibitem{wang2022generalizingSurveyJingdong}
J.~Wang, C.~Lan, C.~Liu, Y.~Ouyang, T.~Qin, W.~Lu, Y.~Chen, W.~Zeng, and P.~Yu,
  ``Generalizing to unseen domains: A survey on domain generalization,''
  \emph{IEEE Transactions on Knowledge and Data Engineering}, 2022.

\bibitem{li2021confounderCICF}
X.~Li, Z.~Zhang, G.~Wei, C.~Lan, W.~Zeng, X.~Jin, and Z.~Chen, ``Confounder
  identification-free causal visual feature learning,'' \emph{arXiv preprint
  arXiv:2111.13420}, 2021.

\bibitem{zhang2022GE-ViTs}
C.~Zhang, M.~Zhang, S.~Zhang, D.~Jin, Q.~Zhou, Z.~Cai, H.~Zhao, X.~Liu, and
  Z.~Liu, ``Delving deep into the generalization of vision transformers under
  distribution shifts,'' in \emph{Proceedings of the IEEE/CVF Conference on
  Computer Vision and Pattern Recognition}, 2022, pp. 7277--7286.

\bibitem{ganin2016domain_dann}
Y.~Ganin, E.~Ustinova, H.~Ajakan, P.~Germain, H.~Larochelle, F.~Laviolette,
  M.~Marchand, and V.~Lempitsky, ``Domain-adversarial training of neural
  networks,'' \emph{The journal of machine learning research}, vol.~17, no.~1,
  pp. 2096--2030, 2016.

\bibitem{borgwardt2006integratingMMD}
K.~M. Borgwardt, A.~Gretton, M.~J. Rasch, H.-P. Kriegel, B.~Sch{\"o}lkopf, and
  A.~J. Smola, ``Integrating structured biological data by kernel maximum mean
  discrepancy,'' \emph{Bioinformatics}, vol.~22, no.~14, pp. e49--e57, 2006.

\bibitem{ren2022multi}
C.-X. Ren, Y.-H. Liu, X.-W. Zhang, and K.-K. Huang, ``Multi-source unsupervised
  domain adaptation via pseudo target domain,'' \emph{IEEE Transactions on
  Image Processing}, vol.~31, pp. 2122--2135, 2022.

\bibitem{feng2021complementary}
H.~Feng, M.~Chen, J.~Hu, D.~Shen, H.~Liu, and D.~Cai, ``Complementary pseudo
  labels for unsupervised domain adaptation on person re-identification,''
  \emph{IEEE Transactions on Image Processing}, vol.~30, pp. 2898--2907, 2021.

\bibitem{moon2022multistage}
J.~Moon, D.~Das, and C.~G. Lee, ``A multistage framework with mean subspace
  computation and recursive feedback for online unsupervised domain
  adaptation,'' \emph{IEEE Transactions on Image Processing}, vol.~31, pp.
  4622--4636, 2022.

\bibitem{bai2021hierarchical}
Y.~Bai, C.~Wang, Y.~Lou, J.~Liu, and L.-Y. Duan, ``Hierarchical
  connectivity-centered clustering for unsupervised domain adaptation on person
  re-identification,'' \emph{IEEE Transactions on Image Processing}, vol.~30,
  pp. 6715--6729, 2021.

\bibitem{he2016deepResNet}
K.~He, X.~Zhang, S.~Ren, and J.~Sun, ``Deep residual learning for image
  recognition,'' in \emph{Proceedings of the IEEE conference on computer vision
  and pattern recognition}, 2016, pp. 770--778.

\bibitem{krizhevsky2017imagenetAlexNet}
A.~Krizhevsky, I.~Sutskever, and G.~E. Hinton, ``Imagenet classification with
  deep convolutional neural networks,'' \emph{Communications of the ACM},
  vol.~60, no.~6, pp. 84--90, 2017.

\bibitem{wei2021toalign}
G.~Wei, C.~Lan, W.~Zeng, Z.~Zhang, and Z.~Chen, ``Toalign: Task-oriented
  alignment for unsupervised domain adaptation,'' \emph{Advances in Neural
  Information Processing Systems}, vol.~34, pp. 13\,834--13\,846, 2021.

\bibitem{tzeng2017adversarialADDA}
E.~Tzeng, J.~Hoffman, K.~Saenko, and T.~Darrell, ``Adversarial discriminative
  domain adaptation,'' in \emph{Proceedings of the IEEE conference on computer
  vision and pattern recognition}, 2017, pp. 7167--7176.

\bibitem{saito2017asymmetric}
K.~Saito, Y.~Ushiku, and T.~Harada, ``Asymmetric tri-training for unsupervised
  domain adaptation,'' in \emph{International Conference on Machine
  Learning}.\hskip 1em plus 0.5em minus 0.4em\relax PMLR, 2017, pp. 2988--2997.

\bibitem{zou2019confidenceCRST}
Y.~Zou, Z.~Yu, X.~Liu, B.~Kumar, and J.~Wang, ``Confidence regularized
  self-training,'' in \emph{Proceedings of the IEEE/CVF International
  Conference on Computer Vision}, 2019, pp. 5982--5991.

\bibitem{gu2020spherical}
X.~Gu, J.~Sun, and Z.~Xu, ``Spherical space domain adaptation with robust
  pseudo-label loss,'' in \emph{Proceedings of the IEEE/CVF Conference on
  Computer Vision and Pattern Recognition}, 2020, pp. 9101--9110.

\bibitem{liu2021Swin}
Z.~Liu, Y.~Lin, Y.~Cao, H.~Hu, Y.~Wei, Z.~Zhang, S.~Lin, and B.~Guo, ``Swin
  transformer: Hierarchical vision transformer using shifted windows,'' in
  \emph{Proceedings of the IEEE/CVF International Conference on Computer
  Vision}, 2021, pp. 10\,012--10\,022.

\bibitem{wang2021pyramidPVT}
W.~Wang, E.~Xie, X.~Li, D.-P. Fan, K.~Song, D.~Liang, T.~Lu, P.~Luo, and
  L.~Shao, ``Pyramid vision transformer: A versatile backbone for dense
  prediction without convolutions,'' in \emph{Proceedings of the IEEE/CVF
  International Conference on Computer Vision}, 2021, pp. 568--578.

\bibitem{xu2022groupvit}
J.~Xu, S.~De~Mello, S.~Liu, W.~Byeon, T.~Breuel, J.~Kautz, and X.~Wang,
  ``Groupvit: Semantic segmentation emerges from text supervision,'' in
  \emph{Proceedings of the IEEE/CVF Conference on Computer Vision and Pattern
  Recognition}, 2022, pp. 18\,134--18\,144.

\bibitem{zhu2020deformable}
X.~Zhu, W.~Su, L.~Lu, B.~Li, X.~Wang, and J.~Dai, ``Deformable detr: Deformable
  transformers for end-to-end object detection,'' \emph{arXiv preprint
  arXiv:2010.04159}, 2020.

\bibitem{li2022hst}
B.~Li, X.~Li, Y.~Lu, S.~Liu, R.~Feng, and Z.~Chen, ``Hst: Hierarchical swin
  transformer for compressed image super-resolution,'' \emph{arXiv preprint
  arXiv:2208.09885}, 2022.

\bibitem{liu2022swiniqa}
J.~Liu, X.~Li, Y.~Peng, T.~Yu, and Z.~Chen, ``Swiniqa: Learned swin distance
  for compressed image quality assessment,'' in \emph{Proceedings of the
  IEEE/CVF Conference on Computer Vision and Pattern Recognition}, 2022, pp.
  1795--1799.

\bibitem{lu2022rtn}
Y.~Lu, J.~Fu, X.~Li, W.~Zhou, S.~Liu, X.~Zhang, W.~Wu, C.~Jia, Y.~Liu, and
  Z.~Chen, ``Rtn: Reinforced transformer network for coronary ct angiography
  vessel-level image quality assessment,'' in \emph{International Conference on
  Medical Image Computing and Computer-Assisted Intervention}.\hskip 1em plus
  0.5em minus 0.4em\relax Springer, 2022, pp. 644--653.

\bibitem{xu2021cdtransCDTrans}
T.~Xu, W.~Chen, P.~Wang, F.~Wang, H.~Li, and R.~Jin, ``Cdtrans: Cross-domain
  transformer for unsupervised domain adaptation,'' \emph{arXiv preprint
  arXiv:2109.06165}, 2021.

\bibitem{yang2021tvtTVT}
J.~Yang, J.~Liu, N.~Xu, and J.~Huang, ``Tvt: Transferable vision transformer
  for unsupervised domain adaptation,'' \emph{arXiv preprint arXiv:2108.05988},
  2021.

\bibitem{zhang2022delvingGE-ViTs}
C.~Zhang, M.~Zhang, S.~Zhang, D.~Jin, Q.~Zhou, Z.~Cai, H.~Zhao, X.~Liu, and
  Z.~Liu, ``Delving deep into the generalization of vision transformers under
  distribution shifts,'' in \emph{Proceedings of the IEEE/CVF Conference on
  Computer Vision and Pattern Recognition}, 2022, pp. 7277--7286.

\bibitem{sun2022safeSSRT}
T.~Sun, C.~Lu, T.~Zhang, and H.~Ling, ``Safe self-refinement for
  transformer-based domain adaptation,'' in \emph{Proceedings of the IEEE/CVF
  Conference on Computer Vision and Pattern Recognition}, 2022, pp. 7191--7200.

\bibitem{hoyer2022daformerDAformer}
L.~Hoyer, D.~Dai, and L.~Van~Gool, ``Daformer: Improving network architectures
  and training strategies for domain-adaptive semantic segmentation,'' in
  \emph{Proceedings of the IEEE/CVF Conference on Computer Vision and Pattern
  Recognition}, 2022, pp. 9924--9935.

\bibitem{wang2022domainBCAT}
X.~Wang, P.~Guo, and Y.~Zhang, ``Domain adaptation via bidirectional
  cross-attention transformer,'' \emph{arXiv preprint arXiv:2201.05887}, 2022.

\bibitem{vaswani2017attentionisallyouneed}
A.~Vaswani, N.~Shazeer, N.~Parmar, J.~Uszkoreit, L.~Jones, A.~N. Gomez,
  {\L}.~Kaiser, and I.~Polosukhin, ``Attention is all you need,''
  \emph{Advances in neural information processing systems}, vol.~30, 2017.

\bibitem{jang2016categoricalGumbelsoftmax}
E.~Jang, S.~Gu, and B.~Poole, ``Categorical reparameterization with
  gumbel-softmax,'' \emph{arXiv preprint arXiv:1611.01144}, 2016.

\bibitem{peng2019momentDomainNet}
X.~Peng, Q.~Bai, X.~Xia, Z.~Huang, K.~Saenko, and B.~Wang, ``Moment matching
  for multi-source domain adaptation,'' in \emph{Proceedings of the IEEE/CVF
  international conference on computer vision}, 2019, pp. 1406--1415.

\bibitem{venkateswara2017deepOffice-Home}
H.~Venkateswara, J.~Eusebio, S.~Chakraborty, and S.~Panchanathan, ``Deep
  hashing network for unsupervised domain adaptation,'' in \emph{Proceedings of
  the IEEE conference on computer vision and pattern recognition}, 2017, pp.
  5018--5027.

\bibitem{peng2017visdaViSDA2017}
X.~Peng, B.~Usman, N.~Kaushik, J.~Hoffman, D.~Wang, and K.~Saenko, ``Visda: The
  visual domain adaptation challenge,'' \emph{arXiv preprint arXiv:1710.06924},
  2017.

\bibitem{deng2021joint}
W.~Deng, Q.~Liao, L.~Zhao, D.~Guo, G.~Kuang, D.~Hu, and L.~Liu, ``Joint
  clustering and discriminative feature alignment for unsupervised domain
  adaptation,'' \emph{IEEE Transactions on Image Processing}, vol.~30, pp.
  7842--7855, 2021.

\bibitem{xu2021neutral}
H.~Xu, M.~Yang, L.~Deng, Y.~Qian, and C.~Wang, ``Neutral cross-entropy loss
  based unsupervised domain adaptation for semantic segmentation,'' \emph{IEEE
  Transactions on Image Processing}, vol.~30, pp. 4516--4525, 2021.

\bibitem{xu2022few}
B.~Xu, Z.~Zeng, C.~Lian, and Z.~Ding, ``Few-shot domain adaptation via mixup
  optimal transport,'' \emph{IEEE Transactions on Image Processing}, vol.~31,
  pp. 2518--2528, 2022.

\bibitem{dai2021disentangling}
P.~Dai, P.~Chen, Q.~Wu, X.~Hong, Q.~Ye, Q.~Tian, C.-W. Lin, and R.~Ji,
  ``Disentangling task-oriented representations for unsupervised domain
  adaptation,'' \emph{IEEE Transactions on Image Processing}, vol.~31, pp.
  1012--1026, 2021.

\bibitem{lu2022styleam}
Y.~Lu, X.~Li, J.~Liu, and Z.~Chen, ``Styleam: Perception-oriented unsupervised
  domain adaption for non-reference image quality assessment,'' \emph{arXiv
  preprint arXiv:2207.14489}, 2022.

\bibitem{liu2022source}
J.~Liu, X.~Li, S.~An, and Z.~Chen, ``Source-free unsupervised domain adaptation
  for blind image quality assessment,'' \emph{arXiv preprint arXiv:2207.08124},
  2022.

\bibitem{zellinger2017centralCMD}
W.~Zellinger, T.~Grubinger, E.~Lughofer, T.~Natschl{\"a}ger, and
  S.~Saminger-Platz, ``Central moment discrepancy (cmd) for domain-invariant
  representation learning,'' \emph{arXiv preprint arXiv:1702.08811}, 2017.

\bibitem{sun2016returnCORAL}
B.~Sun, J.~Feng, and K.~Saenko, ``Return of frustratingly easy domain
  adaptation,'' in \emph{Proceedings of the AAAI Conference on Artificial
  Intelligence}, vol.~30, no.~1, 2016.

\bibitem{sun2016deepdeepcoral}
B.~Sun and K.~Saenko, ``Deep coral: Correlation alignment for deep domain
  adaptation,'' in \emph{European conference on computer vision}.\hskip 1em
  plus 0.5em minus 0.4em\relax Springer, 2016, pp. 443--450.

\bibitem{peng2019momentM3SDA}
X.~Peng, Q.~Bai, X.~Xia, Z.~Huang, K.~Saenko, and B.~Wang, ``Moment matching
  for multi-source domain adaptation,'' in \emph{Proceedings of the IEEE/CVF
  international conference on computer vision}, 2019, pp. 1406--1415.

\bibitem{zhang2019domainSymnets}
Y.~Zhang, H.~Tang, K.~Jia, and M.~Tan, ``Domain-symmetric networks for
  adversarial domain adaptation,'' in \emph{Proceedings of the IEEE/CVF
  conference on computer vision and pattern recognition}, 2019, pp. 5031--5040.

\bibitem{hoffman2018cycada}
J.~Hoffman, E.~Tzeng, T.~Park, J.-Y. Zhu, P.~Isola, K.~Saenko, A.~Efros, and
  T.~Darrell, ``Cycada: Cycle-consistent adversarial domain adaptation,'' in
  \emph{International conference on machine learning}.\hskip 1em plus 0.5em
  minus 0.4em\relax Pmlr, 2018, pp. 1989--1998.

\bibitem{russo2018sourceSBADA}
P.~Russo, F.~M. Carlucci, T.~Tommasi, and B.~Caputo, ``From source to target
  and back: symmetric bi-directional adaptive gan,'' in \emph{Proceedings of
  the IEEE Conference on Computer Vision and Pattern Recognition}, 2018, pp.
  8099--8108.

\bibitem{long2018conditionalCDAN}
M.~Long, Z.~Cao, J.~Wang, and M.~I. Jordan, ``Conditional adversarial domain
  adaptation,'' \emph{Advances in neural information processing systems},
  vol.~31, 2018.

\bibitem{saito2018maximumMCD}
K.~Saito, K.~Watanabe, Y.~Ushiku, and T.~Harada, ``Maximum classifier
  discrepancy for unsupervised domain adaptation,'' in \emph{Proceedings of the
  IEEE conference on computer vision and pattern recognition}, 2018, pp.
  3723--3732.

\bibitem{xie2018learningMSTN}
S.~Xie, Z.~Zheng, L.~Chen, and C.~Chen, ``Learning semantic representations for
  unsupervised domain adaptation,'' in \emph{International conference on
  machine learning}.\hskip 1em plus 0.5em minus 0.4em\relax PMLR, 2018, pp.
  5423--5432.

\bibitem{wang2019transferableTADA}
X.~Wang, L.~Li, W.~Ye, M.~Long, and J.~Wang, ``Transferable attention for
  domain adaptation,'' in \emph{Proceedings of the AAAI Conference on
  Artificial Intelligence}, vol.~33, no.~01, 2019, pp. 5345--5352.

\bibitem{chen2022reusingDALN}
L.~Chen, H.~Chen, Z.~Wei, X.~Jin, X.~Tan, Y.~Jin, and E.~Chen, ``Reusing the
  task-specific classifier as a discriminator: Discriminator-free adversarial
  domain adaptation,'' in \emph{Proceedings of the IEEE/CVF Conference on
  Computer Vision and Pattern Recognition}, 2022, pp. 7181--7190.

\bibitem{chen2019progressivePFAN}
C.~Chen, W.~Xie, W.~Huang, Y.~Rong, X.~Ding, Y.~Huang, T.~Xu, and J.~Huang,
  ``Progressive feature alignment for unsupervised domain adaptation,'' in
  \emph{Proceedings of the IEEE/CVF conference on computer vision and pattern
  recognition}, 2019, pp. 627--636.

\bibitem{li2021biBCDM}
S.~Li, F.~Lv, B.~Xie, C.~H. Liu, J.~Liang, and C.~Qin, ``Bi-classifier
  determinacy maximization for unsupervised domain adaptation.'' in
  \emph{AAAI}, vol.~2, 2021, p.~5.

\bibitem{luo2020unsupervised}
Y.-W. Luo, C.-X. Ren, P.~Ge, K.-K. Huang, and Y.-F. Yu, ``Unsupervised domain
  adaptation via discriminative manifold embedding and alignment,'' in
  \emph{Proceedings of the AAAI Conference on Artificial Intelligence},
  vol.~34, no.~04, 2020, pp. 5029--5036.

\bibitem{chang2019domainDSBN}
W.-G. Chang, T.~You, S.~Seo, S.~Kwak, and B.~Han, ``Domain-specific batch
  normalization for unsupervised domain adaptation,'' in \emph{Proceedings of
  the IEEE/CVF conference on Computer Vision and Pattern Recognition}, 2019,
  pp. 7354--7362.

\bibitem{wei2021metaalign}
G.~Wei, C.~Lan, W.~Zeng, and Z.~Chen, ``Metaalign: Coordinating domain
  alignment and classification for unsupervised domain adaptation,'' in
  \emph{Proceedings of the IEEE/CVF Conference on Computer Vision and Pattern
  Recognition}, 2021, pp. 16\,643--16\,653.

\bibitem{finn2017modelMAML}
C.~Finn, P.~Abbeel, and S.~Levine, ``Model-agnostic meta-learning for fast
  adaptation of deep networks,'' in \emph{International conference on machine
  learning}.\hskip 1em plus 0.5em minus 0.4em\relax PMLR, 2017, pp. 1126--1135.

\bibitem{french2017self}
G.~French, M.~Mackiewicz, and M.~Fisher, ``Self-ensembling for visual domain
  adaptation,'' \emph{arXiv preprint arXiv:1706.05208}, 2017.

\bibitem{zou2018unsupervisedCBST}
Y.~Zou, Z.~Yu, B.~Kumar, and J.~Wang, ``Unsupervised domain adaptation for
  semantic segmentation via class-balanced self-training,'' in
  \emph{Proceedings of the European conference on computer vision (ECCV)},
  2018, pp. 289--305.

\bibitem{xie2018learning}
S.~Xie, Z.~Zheng, L.~Chen, and C.~Chen, ``Learning semantic representations for
  unsupervised domain adaptation,'' in \emph{International conference on
  machine learning}.\hskip 1em plus 0.5em minus 0.4em\relax PMLR, 2018, pp.
  5423--5432.

\bibitem{mei2020instance}
K.~Mei, C.~Zhu, J.~Zou, and S.~Zhang, ``Instance adaptive self-training for
  unsupervised domain adaptation,'' in \emph{European conference on computer
  vision}.\hskip 1em plus 0.5em minus 0.4em\relax Springer, 2020, pp. 415--430.

\bibitem{liu2021cycle}
H.~Liu, J.~Wang, and M.~Long, ``Cycle self-training for domain adaptation,''
  \emph{Advances in Neural Information Processing Systems}, vol.~34, pp.
  22\,968--22\,981, 2021.

\bibitem{na2021fixbi}
J.~Na, H.~Jung, H.~J. Chang, and W.~Hwang, ``Fixbi: Bridging domain spaces for
  unsupervised domain adaptation,'' in \emph{Proceedings of the IEEE/CVF
  Conference on Computer Vision and Pattern Recognition}, 2021, pp. 1094--1103.

\bibitem{zhang2022udaCAUDA}
C.~Zhang and G.~H. Lee, ``Ca-uda: Class-aware unsupervised domain adaptation
  with optimal assignment and pseudo-label refinement,'' \emph{arXiv preprint
  arXiv:2205.13579}, 2022.

\bibitem{zhang2022low}
Y.~Zhang, J.~Li, and Z.~Wang, ``Low-confidence samples matter for domain
  adaptation,'' \emph{arXiv preprint arXiv:2202.02802}, 2022.

\bibitem{zheng2021rethinking}
S.~Zheng, J.~Lu, H.~Zhao, X.~Zhu, Z.~Luo, Y.~Wang, Y.~Fu, J.~Feng, T.~Xiang,
  P.~H. Torr \emph{et~al.}, ``Rethinking semantic segmentation from a
  sequence-to-sequence perspective with transformers,'' in \emph{Proceedings of
  the IEEE/CVF conference on computer vision and pattern recognition}, 2021,
  pp. 6881--6890.

\bibitem{touvron2021trainingDEiT}
H.~Touvron, M.~Cord, M.~Douze, F.~Massa, A.~Sablayrolles, and H.~J{\'e}gou,
  ``Training data-efficient image transformers \& distillation through
  attention,'' in \emph{International Conference on Machine Learning}.\hskip
  1em plus 0.5em minus 0.4em\relax PMLR, 2021, pp. 10\,347--10\,357.

\bibitem{ma2021exploitingWinTR}
W.~Ma, J.~Zhang, S.~Li, C.~H. Liu, Y.~Wang, and W.~Li, ``Exploiting both
  domain-specific and invariant knowledge via a win-win transformer for
  unsupervised domain adaptation,'' \emph{arXiv preprint arXiv:2111.12941},
  2021.

\bibitem{chen2019progressive}
C.~Chen, W.~Xie, W.~Huang, Y.~Rong, X.~Ding, Y.~Huang, T.~Xu, and J.~Huang,
  ``Progressive feature alignment for unsupervised domain adaptation,'' in
  \emph{Proceedings of the IEEE/CVF conference on computer vision and pattern
  recognition}, 2019, pp. 627--636.

\bibitem{xu2019largerSAFN}
R.~Xu, G.~Li, J.~Yang, and L.~Lin, ``Larger norm more transferable: An adaptive
  feature norm approach for unsupervised domain adaptation,'' in
  \emph{Proceedings of the IEEE/CVF International Conference on Computer
  Vision}, 2019, pp. 1426--1435.

\bibitem{lee2019slicedSWD}
C.-Y. Lee, T.~Batra, M.~H. Baig, and D.~Ulbricht, ``Sliced wasserstein
  discrepancy for unsupervised domain adaptation,'' in \emph{Proceedings of the
  IEEE/CVF Conference on Computer Vision and Pattern Recognition}, 2019, pp.
  10\,285--10\,295.

\bibitem{liang2020weSHOT}
J.~Liang, D.~Hu, and J.~Feng, ``Do we really need to access the source data?
  source hypothesis transfer for unsupervised domain adaptation,'' in
  \emph{International Conference on Machine Learning}.\hskip 1em plus 0.5em
  minus 0.4em\relax PMLR, 2020, pp. 6028--6039.

\bibitem{van2017neuralVQVAE}
A.~Van Den~Oord, O.~Vinyals \emph{et~al.}, ``Neural discrete representation
  learning,'' \emph{Advances in neural information processing systems},
  vol.~30, 2017.

\bibitem{wang2019transferableCDAN+TN}
X.~Wang, Y.~Jin, M.~Long, J.~Wang, and M.~I. Jordan, ``Transferable
  normalization: Towards improving transferability of deep neural networks,''
  \emph{Advances in neural information processing systems}, vol.~32, 2019.

\bibitem{li2021semanticDCAN+SCDA}
S.~Li, M.~Xie, F.~Lv, C.~H. Liu, J.~Liang, C.~Qin, and W.~Li, ``Semantic
  concentration for domain adaptation,'' in \emph{Proceedings of the IEEE/CVF
  International Conference on Computer Vision}, 2021, pp. 9102--9111.

\bibitem{han2022learning}
Z.~Han, H.~Sun, and Y.~Yin, ``Learning transferable parameters for unsupervised
  domain adaptation,'' \emph{IEEE Transactions on Image Processing}, vol.~31,
  pp. 6424--6439, 2022.

\end{thebibliography}


\end{document}